%% file: acl_latex.tex
\pdfoutput=1
\documentclass[11pt]{article}

\usepackage[preprint]{acl}

\usepackage{times}
\usepackage{latexsym}

\usepackage[T1]{fontenc}
\usepackage[utf8]{inputenc}

\usepackage{enumitem}
\usepackage{microtype}
\usepackage{amsmath}
\usepackage{amssymb}
\usepackage{algorithm}
\usepackage{algpseudocode}
\usepackage{inconsolata}
\usepackage{soul}
\usepackage{xcolor}

\usepackage{graphicx}
\usepackage{tabularx}
\usepackage{subfigure}
\usepackage{booktabs}   % for \toprule, \midrule, \bottomrule
\usepackage{multirow}   % for \multirow
\usepackage{colortbl}   % for \rowcolor
\usepackage{xcolor}     % for gray color in \rowcolor
\newcommand \footnoteONLYtext[1]
{
	\let \mybackup \thefootnote
	\let \thefootnote \relax
	\footnotetext{#1}
	\let \thefootnote \mybackup
	\let \mybackup \imareallyundefinedcommand
}
\title{AttnComp: Attention-Guided Adaptive Context Compression \\for Retrieval-Augmented Generation}

\author{
  \textbf{Lvzhou Luo\textsuperscript{1,2,3}},
  \textbf{Yixuan Cao\textsuperscript{1,2,3*}},
  \textbf{Ping Luo\textsuperscript{1,2,3*}}
\\
  \textsuperscript{1}Key Lab of Intelligent Information Processing, Institute of Computing Technology,\\ Chinese Academy of Sciences (CAS), Beijing 100190, China
\\
  \textsuperscript{2}State Key Lab of AI Safety, Beijing 100190, China
\\
  \textsuperscript{3}University of Chinese Academy of Sciences, CAS, Beijing 100049, China
\\
 \texttt{\{luolvzhou23s,caoyixuan,luop\}@ict.ac.cn}
}

\begin{document}
\maketitle
\begin{abstract}

  Retrieval-augmented generation improves the factual accuracy of Large Language Models (LLMs) by incorporating external context, but often suffers from irrelevant retrieved content that hinders effectiveness. Context compression addresses this issue by filtering out irrelevant information from context before LLM generation. However, existing methods struggle to adaptively adjust compression rates for different context, maintain low latency and integrate information across multiple documents. To overcome these limitations, We introduce AttnComp, an adaptive, efficient and context-aware compression framework. By leveraging the attention mechanism of LLMs to identify relevant information, AttnComp employs a Top-P compression algorithm to retain the minimal set of documents whose cumulative attention weights exceeds a predefined threshold. In addition to compression, AttnComp estimates response confidence by assessing the overall relevance of the retrieved content, enabling users to gauge response reliability. Experiments demonstrate that AttnComp outperforms existing compression methods and uncompressed baselines, achieving higher accuracy with substantial compression rates and lower latency.

\end{abstract}

\footnoteONLYtext{\textsuperscript{*}Corresponding author: Yixuan Cao and Ping Luo.}

\input{introduction}
\input{related_work}
\input{observation}
\input{method}
\input{experiments}
\input{analysis}

\section{Conclusion}
We introduce AttnComp, a novel framework that leverages the attention mechanism to adaptively compress retrieved documents. Additionally, AttnComp provides a confidence estimation capability for evaluating RAG responses. Extensive experiments demonstrate that AttnComp outperforms existing compression methods and uncompressed baseline, offering higher accuracy with significant compression rates and lower end-to-end latency.

\section{Limitations}
Our study has several limitations. First, all observations and experiments are conducted on LLMs with up to 8 billion parameters due to computational constraints, and we do not evaluate the effectiveness of our method on larger models. Investigating AttnComp's performance across a broader range of model sizes may yield valuable insights. Second, the focus of this work is on the attention mechanisms of dense model architectures, leaving the applicability of our approach to other architectures, such as Mixture-of-Experts (MoE) models, unexplored. Third, our automated annotation strategies relied on Llama-3.1-8B-Instruct for data validation. Given the potential for hallucinations in LLMs, some errors may still exist in the constructed dataset. Finally, although the quality of retrieved content is critical for answer generation in RAG systems, other factors—such as the inherent parameter knowledge in the LLM and the way it integrates retrieved information—also affect response quality\citep{chen2024controlling}. Our proposed confidence estimation method focuses solely on the quality of retrieved documents, which may lead to inaccurate assessments when other influential factors are at play.

\section*{Ethics Statement}
Our work improves the accuracy and efficiency of RAG systems by compressing retrieved content before generation. However, the proposed compression model may exhibit topic-dependent preferences, leading to unequal treatment of information and potential biases in the outputs. These issues raise concerns about fairness and reliability, especially in sensitive or high-stakes domains. Future work should mitigate such risks through fairness-aware training and bias detection to ensure ethical deployment.

\section*{Acknowledgments}
The work was supported by National Key Research and Development Program of China (No.2022YFB2702502), the National Natural Science Foundation of China (No.62206265, 62076231).

\bibliography{custom}

\appendix
\input{appendix}

\end{document}

%% file: introduction.tex
\section{Introduction}

\begin{figure*}[t]
    \centering
    \includegraphics[width=\linewidth]{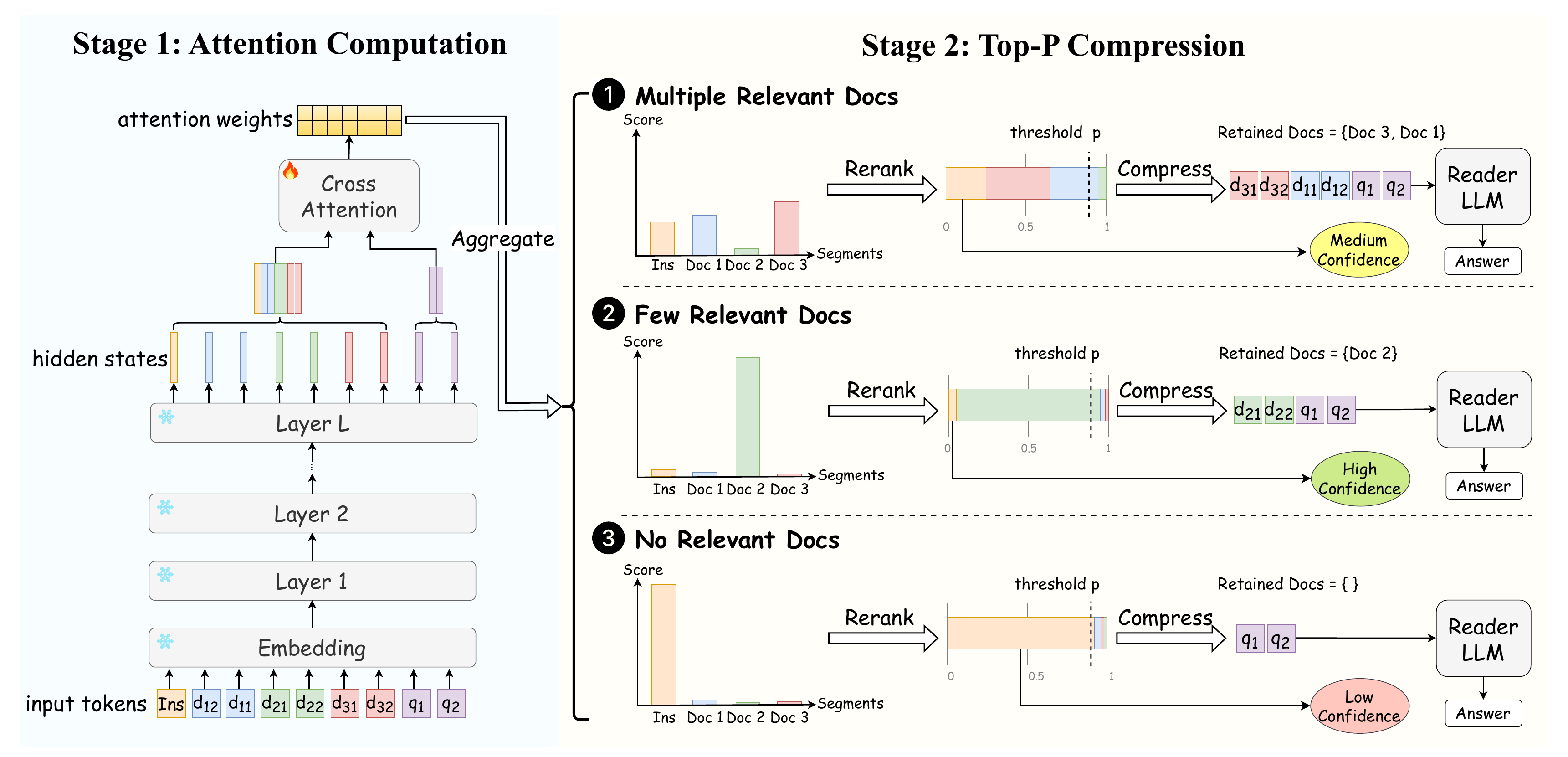}
    \caption{Illustration of the AttnComp Framework. AttnComp consists of two stages: Stage 1 involves attention computation, where attention weights are calculated from the query (q) to the context composed of the instruction (Ins) and documents (d); Stage 2 applies a Top-P compression algorithm to select the most relevant documents and generate a confidence score for the RAG response. Three cases are illustrated: (1) Documents 1 and 3 are relevant and retained; (2) Only Document 2 is relevant and retained; (3) All documents are irrelevant and filtered out.}
    \label{fig:overview}
\end{figure*}

Retrieval-Augmented Generation (RAG) enhances the factual accuracy and reliability of Large Language Models (LLMs) in knowledge-intensive tasks by integrating retrieved context into their generation process\citep{lewis2020retrieval, borgeaud2022improving, izacard2023atlas, ram2023context, xu2023retrieval}. However, practical RAG applications often grapple with retrieved content containing substantial irrelevant information, even entirely unrelated to the query\citep{sauchuk2022role}. This gives rise to three primary issues: first, LLMs can be misled by such noise, leading to incorrect answers\citep{shi2023large, jin2024long, yoran2024making, wu2024easily}; second, LLMs struggle to identify and utilize key information effectively as context length increases\citep{liu2024lost}; and third, irrelevant content unnecessarily inflates input sequences, escalating computational overhead.

To mitigate these issues, context compression has emerged as a promising solution to filter out irrelevant information before generation. Existing methods can be categorized into abstractive and extractive approaches. Abstractive methods leverage LLMs to summarize or rewrite retrieved content via autoregressive generation\citep{xu2023recomp, yoon2024compact, zhu2024information}. While achieving high compression rates, they incur significant latency due to token-by-token decoding. Extractive methods instead select relevant spans from the original content, offering greater efficiency\citep{jiang2024longllmlingua, hwang2024exit, chirkova2025provence}. However, current extractive methods typically only assess the relevance of individual sentence or document to the query, limiting their ability to integrate information across broader context. Furthermore, many such approaches rely on fixed compression rates or target lengths\citep{xu2023recomp, jiang2024longllmlingua}, ignoring the variable proportion of relevant content and risking under- or over-compression.

Consequently, we posit that an effective context compression method should exhibit three key properties: (1) \textbf{Adaptive}: It should dynamically adjust the compression rates based on the proportion of relevant information within the context.  (2) \textbf{Efficient}: It should maintain low computational cost and latency, ensuring rapid processing for real-time applications. (3) \textbf{Context-Aware}: It should integrate and synthesize information from the entire retrieved content to accurately identify relevant segments.  However, to the best of our knowledge, no existing compression method simultaneously satisfies all three of these properties.

To bridge this gap, we introduce \textit{AttnComp} (Attention-guided Context Compression), an adaptive, efficient and context-aware extractive compression method that leverages the inherent attention mechanisms of LLMs. As illustrated in Figure~\ref{fig:overview}, the AttnComp pipeline consists of two stages: (1) Attention Computation. Given a prompt that combines the instruction, retrieved documents and query, we compute attention weights from middle layers of the LLM to quantify the relevance of each text segment to the query. (see Sec. \ref{sec: model} for details). (2) Top-P Compression. We aggregate the attention weights to compute scores for the instruction and each document. Documents are then ranked by score, and the top ones are retained until their cumulative score, combined with that of the instruction, reaches a predefined threshold. Compared to fixed-length compression, this approach adjusts the retained content based on attention distribution, allowing for flexible selection ranging from no documents to all documents (see Sec. \ref{sec: top-p} for details).

We observe that while the attention mechanisms in LLMs inherently capture relevance, they can still assign high attention to irrelevant content.  This is particularly evident when all retrieved documents are irrelevant, as the model fails to shift attention away from the documents, with some irrelevant ones consistently receiving high attention. To address this, we fine-tune the cross-attention layer of the model to direct attention to relevant documents when present, or to the instruction when all documents are irrelevant (see Sec. \ref{sec: finetune} for details). 

Experimental results on multiple QA datasets highlight the superior performance of AttnComp. It achieves a 1.9 point accuracy improvement over the uncompressed baseline, while other compression methods incur at least a 3 point decrease. This advantage is even more pronounced in multi-hop question answering, which requires integrating information from multiple documents and thus places higher demands on context-aware compression capabilities. Here, our method yields at least a 5.4 point improvement over other sentence-level compression methods. Beyond accuracy, AttnComp achieves a 17x compression rate, outperforming all other evaluated extractive methods, and significantly reduces the RAG system's end-to-end latency to 49\% of the uncompressed baseline.

Beyond its primary role in compression, AttnComp also offers a valuable capability for estimating the confidence of RAG responses by leveraging the attention assigned to the instruction (see Sec. \ref{sec: confidence} for details). After training, the attention allocated to the instruction correlates with the quality of retrieved documents, serving as an indicator of answer reliability. Experiments show a strong positive correlation between the confidence score and actual answer accuracy, enabling users to assess response trustworthiness and mitigate risks from low-quality retrieval. Furthermore, this capability also suggests a possible avenue for future research on autonomous iterative RAG\citep{asai2023self, su2024dragin, yu2024auto}.

In summary, our contributions are as follows:

1. We propose AttnComp, a novel extractive compression framework for RAG that is adaptive, efficient, and context-aware.

2. Our method enables confidence estimation for RAG responses, allowing users assess reliability and mitigate risks from low-quality retrieval.

3. Extensive experiments show that AttnComp outperforms existing compression methods and uncompressed retrieval baselines, delivering higher accuracy and lower end-to-end latency.

%% file: related_work.tex
\section{Related Work}

\noindent \textbf{Context Compression.} The compress methods can be broadly categorized into abstractive and extractive approaches. For abstractive compression, RECOMP-abs \citep{xu2023recomp} trains a T5-based model to summarize the retrieved content. \citet{zhu2024information} leverage the Information Bottleneck principle to train LLMs for summarization. CompAct~\citep{yoon2024compact} employs LLMs to summarize retrieved passages and introduces an iterative strategy that progressively updates the relevant context as new passages are incorporated. For extractive compression, RECOMP-ext \citep{xu2023recomp} performs sentence-level semantic matching by selecting the top-k sentences whose embeddings are most similar to the query. LongLLMLingua\citep{jiang2024longllmlingua} proposes a perplexity-based metric to assess the relevance between context and question. A critical limitation of these methods is their dependence on fixed compression ratios. To allow more flexible and adaptive compression, EXIT \citep{hwang2024exit} employs LLMs to conduct binary relevance classification for each sentence, enabling adaptive context reduction. Provence \citep{chirkova2025provence} trains a lightweight DeBERTa model\citep{he2021debertav3} to predict sentence-level relevance scores and retains the sentences that exceed a predefined threshold.

\noindent \textbf{Confidence Estimation.}  Estimating model confidence helps mitigate the risk of unreliable outputs from LLMs \citep{geng2024survey}. Logit-based methods evaluate sentence-level uncertainty using token-level probabilities or entropy \citep{huang2023look, kuhnsemantic2023}. Consistency-based methods estimate confidence by measuring the agreement across multiple generations \citep{manakul2023selfcheckgpt}. However, these approaches focus solely on confidence estimation based on internal knowledge, without considering the integration of external knowledge under the retrieval-augmented generation (RAG) paradigm. \citet{chen2024controlling} highlight two key latent factors influencing confidence in RAG: the quality of the retrieved content and the manner in which it is incorporated into the generation process. To the best of our knowledge, there are currently no methods that estimate confidence in RAG outputs by explicitly evaluating retrieval quality.

%% file: observation.tex
\section{Observations} \label{sec: observations}

\begin{figure*}[]
    \centering
    \subfigure[]{\label{fig:filter-heads}\includegraphics[width=0.48\linewidth]{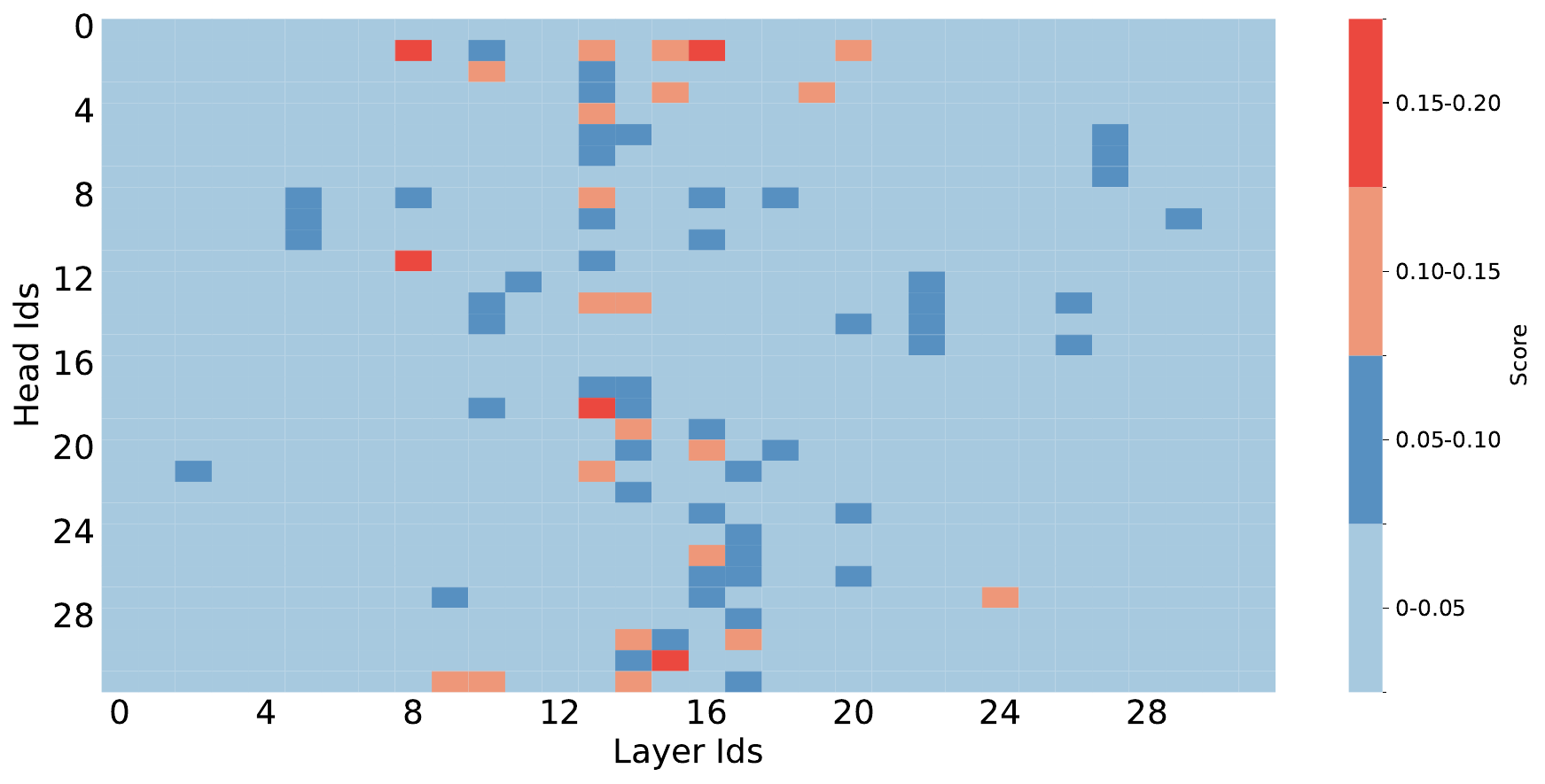}}
    \hfill
    \subfigure[]{\label{fig:adaptive-attention}\includegraphics[width=0.24\linewidth]{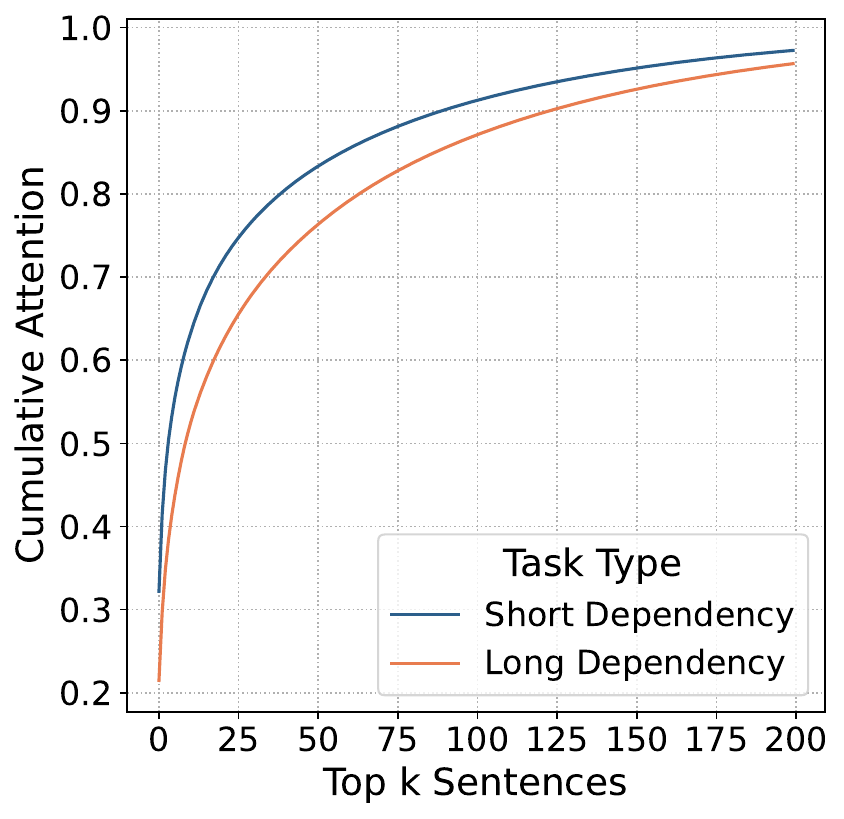}}
    \hfill
    \subfigure[]{\label{fig:attention-sink}\includegraphics[width=0.24\linewidth]{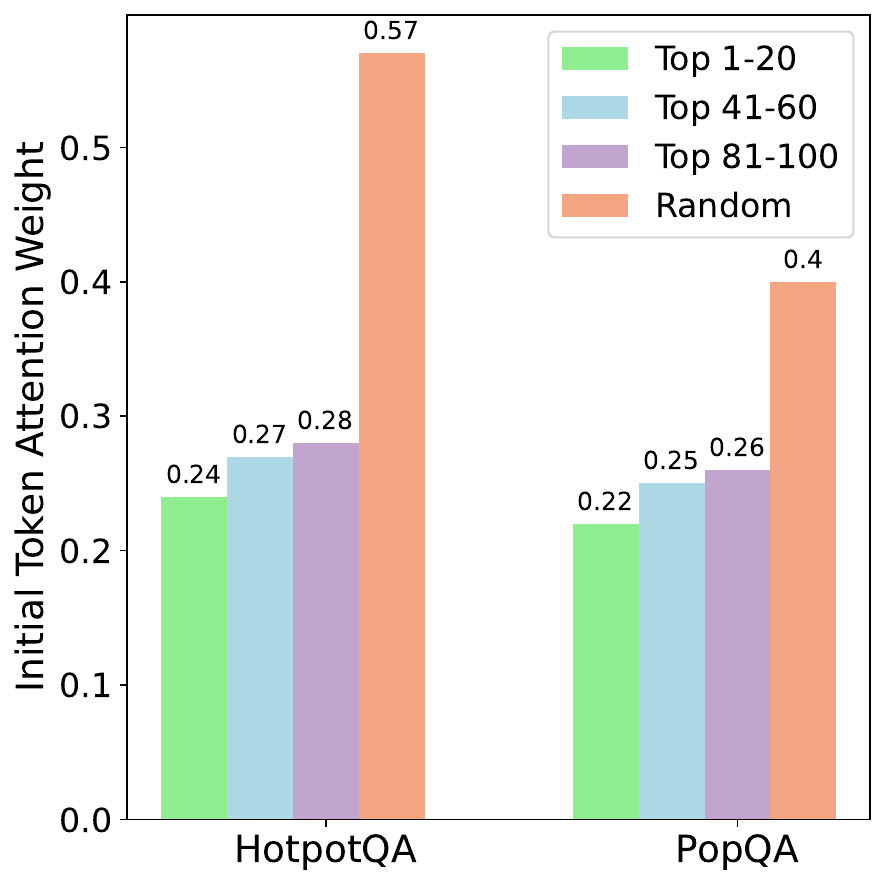}}
    \vspace{-0.4cm}
    \caption{Observations on attention allocation patterns. (a) Attention weights assigned by each attention head to the supporting evidence sentences. (b) Cumulative attention over the top-$k$ most attended sentences in short-dependency and long-dependency tasks. (c) Attention weights on the initial token under different context settings: top 1–20 retrieved documents, top 41–60 documents, top 81–100 documents, and 20 randomly sampled documents.}
    \label{fig:attention-overview}
\end{figure*}

In this section, we present our observations on the attention patterns within LLMs. Our analysis is conducted on QA datasets, where inputs are constructed by concatenating the context before the query. We then compute the attention score from the query to different context segments as follows:
\begin{equation}
    s = \frac{1}{|\mathcal{I}_q|}\sum_{i\in \mathcal{I}_q}\sum_{j\in \mathcal{I}_d} a_{ij}
    \label{eq: attention-score}
\end{equation}
where $\mathcal{I}_q$ and $\mathcal{I}_d$ denote the token indices of the query and context segment, respectively, and $a_{ij}$ is the attention weight from query token $i$ to context token $j$.

We present the experimental details in Appendix \ref{appendix: observations}. The key findings are summarized below:

\begin{itemize}[leftmargin=1em, itemsep=4pt, topsep=2pt, parsep=0pt]

\item \textbf{Certain middle-layer attention heads effectively identify relevant information.} Using the LooGLE benchmark~\citep{li2024loogle}, a QA dataset with labeled evidence sentences, we analyze attention score assigned by each head to the evidence. Figure~\ref{fig:filter-heads} visualizes the attention scores from each head in every LLM layer assigns to evidence sentences. It is observed that some attention heads in the middle layers consistently focus more on supporting evidence, suggesting their ability to capture relevance.

\item \textbf{Attention pattern adapts to the density of relevant content.} The LooGLE benchmark divides tasks into short- and long-dependency types. Short-dependency tasks rely on a single sentence or paragraph, while long-dependency tasks require integrating information across multiple segments. We compute the cumulative attention score over the top-$k$ sentences with the highest attention for both task types. Figure~\ref{fig:adaptive-attention} illustrates how the cumulative attention score changes as the sentence count $k$ varies. As shown, attention is more concentrated in short-dependency tasks and more spread out in long-dependency tasks. Figure~\ref{fig: short-long attention} provides a more intuitive comparison.

\item \textbf{Attention to the initial token of the context increases as context becomes less relevant.} To investigate how attention is allocated when the context is irrelevant, we sample questions from HotpotQA~\citep{yang2018hotpotqa} and PopQA~\citep{mallen2023not}, and construct context settings with varying levels of relevance. We then record the attention assigned to the initial token across these settings. As shown in Figure~\ref{fig:attention-sink}, the attention on the initial token increases as the relevance of the context decreases, consistent with prior findings on attention sinks~\citep{xiao2023efficient}.
\end{itemize}

%% file: method.tex
\section{AttnComp}

Inspired by our observations, we propose a novel compression framework \textit{AttnComp}. In this section, we provide a comprehensive explanation of the framework.

\noindent \textbf{Problem Formulation} Given a query $q$, a RAG system retrieves a set of $k$ documents $D = \{d_1, d_2, \ldots, d_k\}$. A language model $M$ then generates an output $y$ conditioned on the retrieved documents and the query, i.e., $M(y\mid D, q)$. Our objective is to filter irrelevant documents from $D$, yielding a reduced subset $D' \subseteq D$ such that the size of $D'$ is minimized while maintaining or even improving the quality of generated answer $M(y \mid D', q)$.

\subsection{Attention Computation}
\label{sec: model}

% ref to DIFFERENTIAL TRANSFORMER
Building on the finding that attention heads in middle layers of LLMs identify relevant information, our compressor model comprises the first $L$ transformer layers from the original LLM, followed by an additional cross-attention layer. We first construct the context by prefixing a predefined instruction to the concatenated retrieved documents. The context and query are then concatenated and input into the model. After processing through the first $L$ layers, we obtain the hidden states $X_c \in \mathbb{R}^{n \times d_{\text{model}}}$ and $X_q \in \mathbb{R}^{m \times d_{\text{model}}}$ for the context and the query, respectively, where $n$ and $m$ denote their lengths, and $d_{\text{model}}$ is the hidden dimension. The cross-attention layer then computes query-context attention weights $A \in \mathbb{R}^{m \times n}$ as follows:
\begin{equation}
    \begin{aligned}
        Q_i &= X_q \cdot W^Q_i, \quad K_i = X_c \cdot W^K_i, \\
        A &= \frac{1}{H} \sum_{i=1}^H \operatorname{softmax}\left(\frac{Q_i K_i^T}{\sqrt{d_a}}\right)
        \end{aligned}
    \end{equation}
where $H$ denotes the number of attention heads, $W^Q_i, W^K_i \in \mathbb{R}^{d_{\text{model}} \times d_a}$ are the query and key projection matrices for head $i$, and $d_a$ is the dimensionality of each attention head.

\subsection{Top-P Compression}
\label{sec: top-p}
Motivated by our finding that the attention mechanism exhibits adaptive patterns across varying questions and contexts (as shown in Figures~\ref{fig:adaptive-attention} and \ref{fig:attention-sink}), we propose a Top-P compression algorithm that leverages the computed query-context attention weights for adaptive context reduction.

The process commences by calculating attention scores for the instruction ($s_{\text{ins}}$) and each document ($s_{d_i}$), derived from aggregating attention weights $A$, as defined in Equation \ref{eq: attention-score}. These scores are then utilized to dynamically select critical documents. Initially, documents are sorted in descending order of their scores, thereby prioritizing candidates with higher attention. A cumulative sum, initialized with $s_{\text{ins}}$, is subsequently accumulated by incrementally adding the scores of these sorted documents. The selection process continues until either the cumulative score exceeds a predefined threshold $p$, or the current document's score is below a minimum threshold $\epsilon$. Algorithm~\ref{alg:basic-algorithm} provides the pseudo-code for this procedure.

\begin{algorithm}[t]  
    \caption{Top-P Compression Algorithm}  
    \label{alg:basic-algorithm}  
    \begin{algorithmic}[1]  
    \State \textbf{Input:} Instruction score $s_{\text{ins}}$, document scores $\{s_{d_1}, s_{d_2}, \dots, s_{d_k}\}$, top-p threshold $p$, and minimum score threshold $\epsilon$.  
    \State \textbf{Output:} Compressed document set $D'$.
    \State $\{d_{(1)}, \dots, d_{(k)}\} \gets \text{argsort}(\{s_{d_i}\}_{i=1}^k \text{, desc.})$
    \State Initialize $sum \gets s_{\text{ins}}$, $D' \gets  \emptyset$.
    \For{$i = 1$ to $k$}
        \If{$sum \geq p$ \textbf{or} $s_{d_{(i)}} < \epsilon$} 
                \State \textbf{break} %\Comment{Termination conditions}
            \EndIf
        \State $sum \gets sum + s_{d_{(i)}}$
        \State $D' \gets D' \cup \{d_{(i)}\}$
    \EndFor
    \State \textbf{Return:} $D'$.  
    \end{algorithmic}  
    \end{algorithm}

This strategy enables adaptive behavior: when many relevant documents disperse attention, more documents are required for their cumulative attention to reach the threshold $p$. Conversely, if relevant documents are few and attention is concentrated, a smaller subset is sufficient. If all documents are irrelevant, attention focused solely on the instruction can reach the threshold, filtering out all documents.

\subsection{Attention Fine-Tuning}
\label{sec: finetune}

Since certain attention heads can inherently focus on relevant context, we initialize the cross-attention layer using selected attention heads from layer $L+1$ of the LLM. However, empirical results show that the untrained compressor still assign relatively high attention to irrelevant segments, particularly when all documents are irrelevant. To improve relevance discrimination, we fine-tune the model while freezing the first $L$ layers and updating only the cross-attention layer. This lightweight approach updates approximately 0.5\% of the total parameters, reducing training cost while preserving generalization.

\noindent \textbf{Data Construction} We prepare training data where each instance comprises a query $q$, retrieved documents $D = \{d_1, \ldots, d_k\}$, and binary relevance labels $R = \{r_1, \ldots, r_k\}$, with each $r_i \in \{0,1\}$ indicating the relevance of $d_i$ to $q$. Upon examining existing QA datasets, we observe that many contain incomplete relevance annotations, with only a small subset of relevant documents labeled. Directly training on such data yields suboptimal performance, while manual annotation is resource-intensive. To address this, we propose an automated annotation pipeline based on question-answer pairs, comprising two stages: labeling and verification. In the labeling stage, we use an untrained compressor to perform multiple rounds of Top-P compression with different document permutations. Documents that are consistently retained across all rounds are labeled as relevant, while the rest are considered irrelevant. In the verification stage, the query and the labeled relevant documents are provided to an LLM to generate an answer. The annotation is accepted only if the generated answer is correct; otherwise, it is discarded. Furthermore, to enrich our training data, we also construct negative instances where all retrieved documents are irrelevant to the query. A detailed description of this annotation pipeline is provided in Appendix~\ref{appendix: dataset}.

\noindent \textbf{Training} Our training objective incorporates two complementary forms of supervision: document-level and instruction-level.

\noindent\textit{Document-level Supervision}: This component enhances discrimination between relevant and irrelevant documents through binary cross-entropy:
\begin{equation} 
L_{doc} = -\sum_{i=1}^k \left[r_i\log s_{d_i} + (1-r_i)\log(1-s_{d_i})\right]
\end{equation}
\noindent\textit{Instruction-level Supervision}: This component directs attention on the instruction if no documents are relevant, and suppresses it otherwise:
\begin{equation}
L_{ins} = -\Big[ r_\text{ins} \log s_\text{ins} + (1 - r_\text{ins}) \log(1 - s_\text{ins}) \Big]
\end{equation}
where $r_\text{ins}\triangleq\mathbb{I}\big(\sum_{i=1}^k r_i=0\big)$ indicates whether none of the retrieved documents are relevant, with $\mathbb{I}(\cdot)$ representing the indicator function.

The final objective combines both components with a balancing hyperparameter $\lambda$:
\begin{equation} 
L = L_{doc} + \lambda L_{ins}
\end{equation}

\subsection{Confidence Estimation}
\label{sec: confidence}
The fine-tuned model tends to pay more attention to the instruction when the overall relevance of retrieved content is low. We leverage this behavior by using the instruction attention score $s_{\text{ins}}$ as a proxy for retrieval quality. Specifically, a higher $s_{\text{ins}}$ suggests that the retrieved content is less relevant to the query. In such cases, the LLM relies more on its internal knowledge, which can lead to less reliable responses. Motivated by this insight, we define the confidence score $p$ of a RAG response as:
\begin{equation}
    p = 1 - s_{\text{ins}}
    \label{eq:conf}
\end{equation}

%% file: experiments.tex
\section{Experiments}
\begin{table*}[]
  \centering
  \resizebox{1.0\textwidth}{!}{
  \begin{tabular}{l|ccc|ccc|ccc|ccc|ccc|ccc}
  \toprule
    {\multirow{2}{*}{\textbf{Methods}}} &
     \multicolumn{3}{c|}{\textbf{HotpotQA}} &
     \multicolumn{3}{c|}{\textbf{2WikiMQA}} &
     \multicolumn{3}{c|}{\textbf{MuSiQue}} &
     \multicolumn{3}{c|}{\textbf{NQ}} &
     \multicolumn{3}{c|}{\textbf{PopQA}} &
     \multicolumn{3}{c}{\textbf{AVG}} \\ \cmidrule{2-19} 
    {}     &     {\textbf{Comp.}} &   {\textbf{F1}}   &   {\textbf{Acc}}   &   {\textbf{Comp.}} &   {\textbf{F1}}   &   {\textbf{Acc}}  &    {\textbf{Comp.}} &   {\textbf{F1}}   &   {\textbf{Acc}} &    {\textbf{Comp.}} &   {\textbf{F1}}   &   {\textbf{Acc}} &    {\textbf{Comp.}} &   {\textbf{F1}}   &   {\textbf{Acc}} & {\textbf{Comp.}} &   {\textbf{F1}}   &   {\textbf{Acc}} \\
    
  \midrule
  \multicolumn{1}{l}{ \textbf{\textit{No Retrieval}}} \\
  {Direct}     
    & - &  26.5 & 23.6
    & - & 26.2 &  34.1
    & - & 11.4 & 8.8
    & - & 27.1 & 23.6
    & - & 24.1 & 31.3
    & - & 23.1 & 24.3 \\
  \midrule
  \multicolumn{1}{l}{\textbf{\textit{Retrieval without Compression}}} \\
  {All Documents}    
  & {1x}  & {46.3} & {42.7}
  & {1x} & {31.9} & {34.7}
  &  {1x} & {19.3} & {15.5}
  &  {1x}  & \textbf{49.9}   & {53.9}
  & {1x} & \textbf{43.9} & {64.7}
  & 1x & 38.3 & 42.3 \\

  {Top 5 Documents}
  & {18.2x}  & {40.4} & {37.9} 
  & {18.3x}  & {25.7} & {30.4} 
  & {18.4x}  & {16.5} & {13.9}
  & {18.3x} & {49.0} & {54.8} 
  & {18.4x} & {38.1} & {60.9}
  & 18.3x & 33.9 & 39.6 \\
  
  {Top 10 Documents}
  & {9.6x}  & {42.6} & {40.0}
  & {9.6x}  & {28.7} & {31.8}
  & {9.6x}  & {18.4} & {15.5}
  & {9.6x} & {48.3}  & \textbf{55.5}
  & {9.6x} & {39.9}  & {64.4}
  & {9.6x} & 35.6 & 41.4 \\
  % \multicolumn{1}{l}{Raw Document}     &      {1x}  & {29.4}  &  {40.3} & {1x}  & {6.5} & {15.6} & {1x}  & 25.4  & 31.2 &  {1x}  & {39.0} & {51.3}  &  {1x} & {68.9}  & 77.1 \\ 
  \midrule
  % \multicolumn{16}{c}{\textit{Retrieval Augmented Generation}} \\ \midrule
  % \multicolumn{16}{c}{\textit{Compression-based Methods}} \\ \midrule
  \multicolumn{1}{l}{\textbf{\textit{Retrieval with Compression}}} \\
  {RECOMP-ext}
  & {8.0x}  &  {40.4} & {37.5}
  & {8.0x}  &  {27.5} & {30.1}
  & {8.1x}  &  {18.6} & {14.5}
  & {8.4x}  & {47.5}  & {48.7}
  & {9.0x} & {31.3}  & {51.8}
  & 8.3x & 33.1 & 36.5  \\ 
  
  {LongLLMLingua}
  & {9.7x} & {42.5} & {39.1}
  & {9.7x} & {30.2} & {31.9}
  & {9.7x} & {17.2} & {13.8}
  & {9.7x} & {42.6} & {48.1}
  & {9.7x} & {40.1} & {62.1}
  & 9.7x & 34.5 & 39.0 \\ 
  % \midrule
  % \multicolumn{16}{c}{\textit{Adaptive Compressor}} \\ \midrule
  {CompAct} 
  & {80.0x} & {45.1} & {40.2}
  & {82.4x} & {29.2} & {33.2}
  & {71.8x} & {18.0} & {16.5}
  & {84.2x} & {47.5} & {48.7}
  & {98.0x} & 39.8 & 58.4
  & {83.3x} & 35.9 & 39.4 \\
  {Provence} 
  & {10.2x} & {42.5} & {39.8}
  & {10.7x} & {26.8} & {29.3}
  & {8.7x}  & {19.8} & {17.8}
  & {6.8x}  & {41.9} & {50.3}
  & {6.9x} & {34.5} & {58.7}
  & 8.7x & 33.1 & 39.2 \\ 
  % \midrule
  % \multicolumn{1}{l}{AttnComp (w/o SFT)} &  {6.1x} & {46.3} & {45.0}  & {6.3x}  & {31.8} & {35.3} & {5.7x}  & {21.1} & {19.2}   &  {5.8x}  & {47.0} & {55.4} & {7.0x} & {41.0} & \textbf{66.3} \\ %p=0.95
  % \multicolumn{1}{l}{AttnComp (w/o SFT)} &  {9.6x} & {45.8} & {44.1}  & {9.1x}  & {30.5} & {34.0} & {7.7x}  & {23.2} & {21.5}   &  {7.7x}  & {48.1} & {56.4} & {12.7x} & {40.5} & \textbf{65.0} \\ %p=0.6
  \rowcolor{blue!10} {AttnComp (w/o SFT)}
  & {14.1x} & {45.5} & {42.5}
  & {17.0x} & {29.7} & {32.4}
  & {13.8x} & {20.9} & {19.5}
  & {16.1x} & {49.1} & {54.8}
  & {24.0x} & {39.8} & {62.3}
  & 17.0x & 36.8 & 42.3 \\ %p=0.5
  \rowcolor{blue!10} {AttnComp (Ours)} 
  & {12.6x} & \textbf{48.3} & \textbf{45.2}
  & {18.4x} & \textbf{32.9} & \textbf{38.1}
  & {16.3x} & \textbf{21.4} & \textbf{19.6}
  & {13.5x} & {48.0} & {53.0}
  & {23.9x} & {41.3} & \textbf{65.1}
  & 17.0x & \textbf{38.4} & \textbf{44.2} \\ 

  \bottomrule
  \end{tabular}}
  \caption{Main results. We use LLaMA-3.1-8B-Instruct~\citep{grattafiori2024llama} as the reader model and retrieve 100 documents for each query. Since our training data includes only a subset of HotpotQA, we perform zero-shot evaluation on the remaining datasets. Comp. denotes the compression rate, calculated as: $\frac{\text{\# of tokens in retrieved documents}}{\text{\# of tokens in compressed text}}$.}
  \label{table:main}
  \vspace{-0.3cm}
\end{table*}
\subsection{Experimental Setup}
\noindent \textbf{Implementation Details}
We use Llama-3.1-8B-Instruct\citep{grattafiori2024llama} as the backbone architecture for AttnComp, retaining $L=13$ transformer layers and $H=16$ attention heads in the cross-attention layer. To train AttnComp, we construct a training dataset from the HotpotQA training split, consisting of 8,000 examples. Each example includes a question and 100 documents. For 2,000 of these examples, all documents are irrelevant to the question. We train the model with the Adam optimizer\citep{kingma2014adam}, using a learning rate of $2\times10^{-4}$ and a batch size of 8 for 8 epochs. The balancing coefficient $\lambda$ is set to 0.8. During inference, we apply the Top-P Compression algorithm with a threshold of $p=0.95$ and $\epsilon=10^{-2}$. Further information is provided in Appendix \ref{appendix: training}.

\noindent \textbf{Datasets and Retrieval Corpus}
We evaluate AttnComp on both single-hop and multi-hop question answering (QA) benchmarks. For single-hop QA, we use Natural Questions (NQ)\citep{kwiatkowski2019natural} and PopQA\citep{mallen2023not}. For multi-hop QA, we evaluate on HotpotQA \citep{yang2018hotpotqa}, 2WikiMultiHopQA\citep{ho2020constructing} and MuSiQue\citep{trivedi2022musique}. Following \citet{jin2024flashrag}, we use the Wikipedia dump from December 2018 as the retrieval corpus\citep{karpukhin2020dense}, where articles are truncated into non-overlapping documents of 100 words each. For each query, we retrieve the top 100 documents using the E5-base-v2 retriever \citep{wang2022text}.

% We first apply document filtering based on the query-document relevance predicted by Attn-Filter, and then prepend the (full or pruned) relevant documents to the query before inputting them into Llama-3.1-8B-Instruct for answer generation.

% ref to An Information Bottleneck Perspective for Effective Noise Filtering on  Retrieval-Augmented Generation
\subsection{Baseline}
We evaluate AttnComp against several baseline methods. To ensure a fair comparison, all baselines employ Llama-3.1-8B-Instruct\citep{grattafiori2024llama} as the reader model for answer generation, while results using other reader models are presented in Appendix~\ref{appendix: additional-results}. The baselines are as follows: (1) \textit{No Retrieval}: The reader model generates answers directly from the input query, without any retrieved context. (2) \textit{Retrieval without Compression}: All retrieved documents are concatenated and fed to the reader model, serving as an uncompressed baseline. For a more fine-grained comparison, we also report results using only the top-5 and top-10 retrieved documents. (3) \textit{Compression Methods}: We compare AttnComp against four compression methods: RECOMP-ext \citep{xu2023recomp}, LongLLMLingua \citep{jiang2024longllmlingua}, CompAct \citep{yoon2024compact} and Provence \citep{chirkova2025provence}. Additionally, we also compare against AttnComp without fine-tuning. Detailed descriptions of these baselines are provided in Appendix \ref{appendix: baseline}.% setting its threshold $p$ to 0.5 to ensure a fair comparison at a similar compression rate. 

\subsection{Main Results}
  
% We evaluate the performance of AttnComp using three metrics: compression rate (Comp.), F1 score, and accuracy (Acc), with the results presented in Table~\ref{table:main}. We observe that, even without fine-tuning, AttnComp achieves high accuracy, validating the inherent ability of attention mechanisms to identify relevant information. After fine-tuning, AttnComp achieves higher accuracy and compression rate.

% on the three multi-hop QA benchmarks, AttnComp achieves an average accuracy improvement of 3.3 points, alongside an average compression rate of 15.7x. This highlights the negative impact of irrelevant information on answer accuracy and shows that our context compression approach effectively filters out such noise, leading to improved performance. For the two single-hop QA datasets, the accuracy under the compressed setting remains comparable to the uncompressed baseline, despite achieving an average 18.7× compression rate. Considering that our model is trained only on HotpotQA, this result further demonstrates its strong generalization ability.

We evaluate the performance of AttnComp using three metrics: compression rate (Comp.), F1 score, and accuracy (Acc), with the results presented in Table~\ref{table:main}. The results demonstrate that, even without fine-tuning, AttnComp consistently outperforms all compression baselines across all benchmarks in terms of both F1 score and accuracy, while achieving a high compression rate.  Moreover, after fine-tuning, AttnComp further extends its advantage, yielding an average accuracy improvement of 1.9 points over the uncompressed baseline. Notably, AttnComp is the only evaluated method that enhances accuracy, whereas all other compression baselines lead to a decrease of at least 3 points. Furthermore, our method maintains a 17x compression rate, which is higher than that of Provence (8.7x), another adaptive extractive compression method.

% This highlights the negative impact of irrelevant information on answer accuracy and demonstrates that our approach effectively filters out such noise.

%% file: analysis.tex
\section{Analysis}
We evaluate AttnComp for its adaptiveness (Sec. \ref{sec:adaptive}), efficiency (Sec. \ref{sec:efficiency}), context-awareness (Sec. \ref{sec:context-aware}), and robustness (Sec. \ref{sec:robustness}). We also present an ablation study (Sec. \ref{sec:ablation}) and validate the reliability of its confidence estimation (Sec. \ref{sec:confidence}).

\subsection{Adaptive Compression Analysis} \label{sec:adaptive}
\begin{figure}[t]
  \centering
  \setlength{\abovecaptionskip}{0.cm}
  \includegraphics[width=0.98\linewidth]{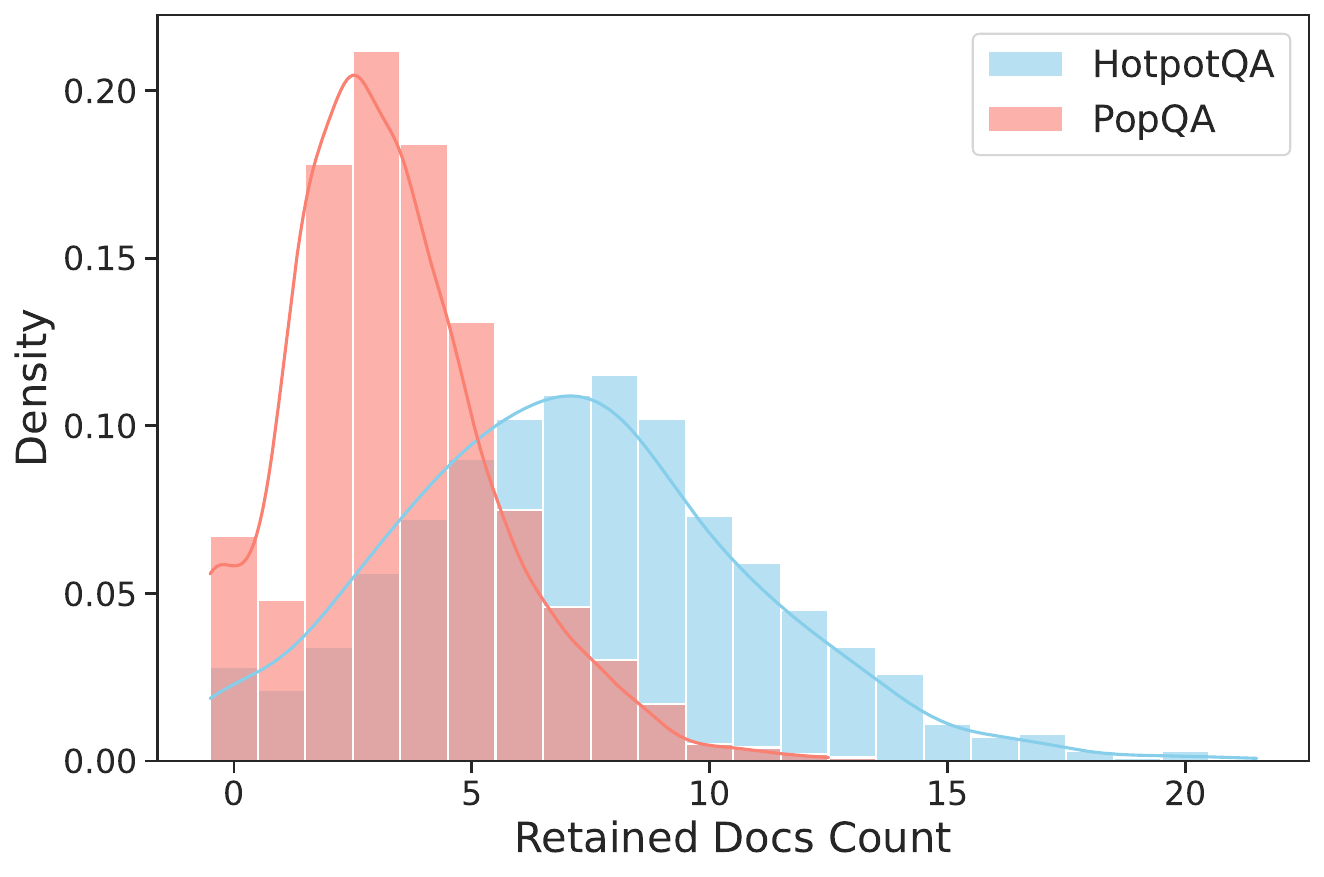}
  \caption{Distribution of documents retained by AttnComp on HotpotQA and PopQA.}
  \label{fig:adaptive}
\end{figure}

To validate the adaptive compression capability of AttnComp, we analyze the number of documents retained after compression on both the HotpotQA and PopQA datasets. As illustrated in Figure \ref{fig:adaptive}, the quantity of retained documents varied dynamically, ranging from 0 to 23. For the multi-hop QA dataset HotpotQA, our model tends to preserve a greater number of documents, averaging 7.5 per query. Conversely, on the simpler PopQA dataset, the number of retained documents was considerably smaller, with an average of 3.7. These results demonstrate that our method can dynamically adjust the compression rate based on the retrieval context and the complexity of the question.

\subsection{Efficiency Analysis} \label{sec:efficiency}

We evaluate the end-to-end latency of the RAG system, including both the compression and generation stages, to demonstrate the efficiency of AttnComp. All methods are tested under the same hardware conditions: one NVIDIA RTX 4090 GPU for compression and two for generation. We report average compression and generation times, excluding retrieval latency as its impact is negligible.

\begin{figure}[t]
  \centering
  \setlength{\abovecaptionskip}{0.cm}
  \includegraphics[width=0.98\linewidth]{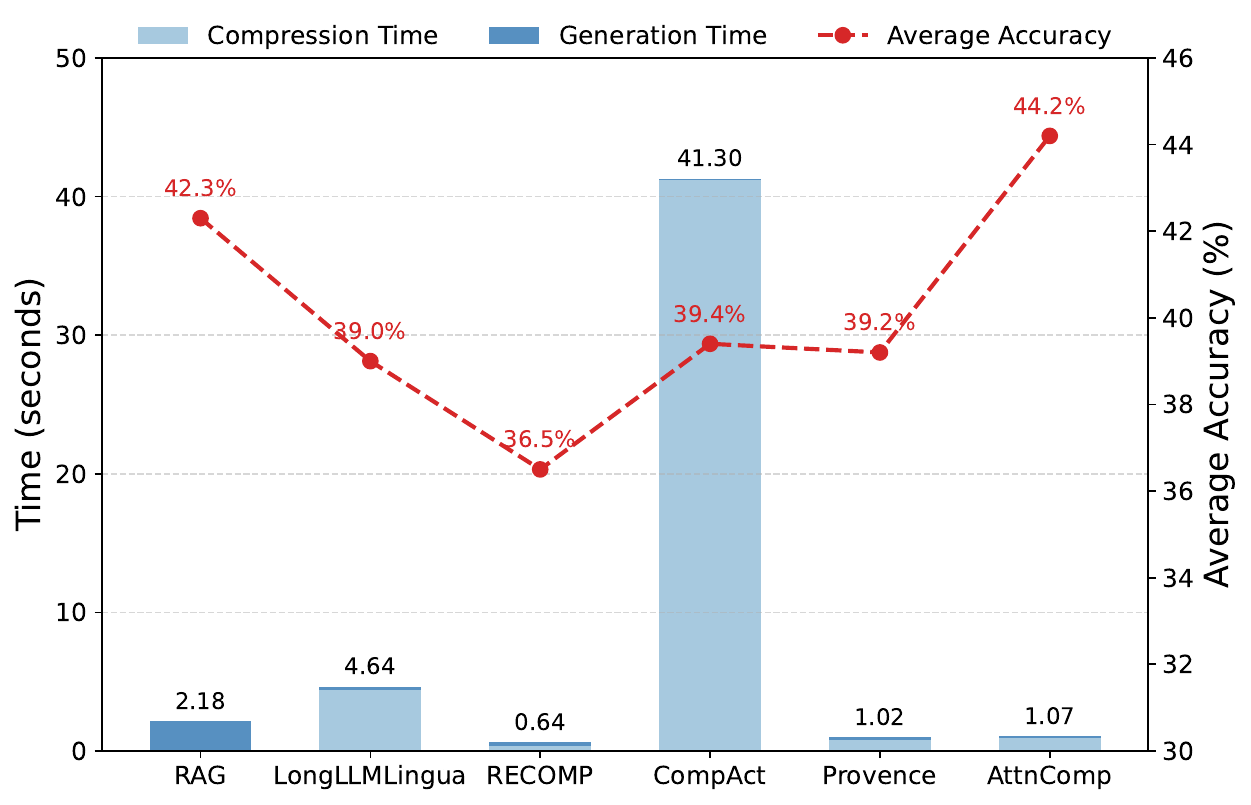}
  \caption{Comparison of end-to-end latency, and average accuracy across baselines and AttnComp.}
  \label{fig:efficiency}
\end{figure}

As illustrated in Figure~\ref{fig:efficiency}, although most compression methods significantly decrease generation time, the compression stage itself introduces considerable latency that cannot be overlooked. For example, while methods like CompAct achieve high compression rates (up to 80x), their reliance on multiple LLM calls during compression incurs substantial latency. This leads to an overall latency (41.30s) that markedly exceeds the uncompressed baseline (2.18s). In contrast, extractive compression methods such as RECOMP and Provence offer lower latency but at the cost of degraded performance. AttnComp achieves efficiency comparable to Provence while delivering better accuracy. With an average compression latency of 0.91 seconds and a generation latency of 0.16 seconds, AttnComp reduces the total end-to-end latency to 49\% of the uncompressed baseline while simultaneously improving answer quality.
\subsection{Context-Aware Compression Analysis} \label{sec:context-aware}
On multi-hop datasets requiring the integration of information from multiple documents, AttnComp exhibits particularly notable improvements, achieving an average accuracy increase of 3.3 points over the uncompressed baseline. Furthermore, on the 2WikiMultiHopQA dataset, it surpasses the sentence-level compression method Provence by a significant 8.8 points in accuracy. This underscores the context-aware capabilities of our approach. A case study is provided in Appendix \ref{appendix: case-study} to demonstrate AttnComp's context-aware capability.

\subsection{Robustness Analysis} \label{sec:robustness}

We evaluate AttnComp across various settings, including different numbers of retrieved documents, top-p thresholds, and context granularities, to demonstrate its effectiveness in diverse scenarios. Experimental details are provided in Appendix \ref{appendix: robustness}, and the main conclusions are as follows:

\noindent \textbf{Varying Number of Retrieved Documents}: We conduct experiments by varying the number of retrieved documents $k$. As shown in Figure~\ref{fig:top-k}, our approach consistently achieves accuracy comparable to or superior to the uncompressed baseline across different values of $k$. Notably, the superiority of our approach over the baseline becomes more substantial as $k$ increases.

\noindent \textbf{Varying Top-p Thresholds}: We evaluate AttnComp with different top-p compression thresholds. As depicted in Figure~\ref{fig:top-p}, the value of $p$ serves as a parameter to balance accuracy against compression rate. Our findings indicate that AttnComp consistently delivers stable and strong performance when $p$ is set to 0.9 or higher.

\noindent \textbf{Varying Context Granularities}: We evaluate AttnComp beyond its standard document-level compression by assessing sentence-level compression performance. Results in Table~\ref{tab:granularity} show that sentence-level compression maintained comparable accuracy to document-level, while achieving a superior compression rate. Additionally, visualizing the attention distribution revealed that, despite being trained with document-level annotations, the model effectively focuses attention on relevant sentences and words, demonstrating its adaptability to different context granularities.

\subsection{Ablation Study} \label{sec:ablation}
\begin{table}[t]
  \centering
  \small
  \begin{tabular}{c|cc|cc}
  \toprule
  \multirow{2}{*}{Arch} & \multicolumn{2}{c|}{w/o Fine-tuning} & \multicolumn{2}{c}{Fine-tuning} \\
  \cmidrule(lr){2-3} \cmidrule(lr){4-5}
                       & Acc   & Comp.    & Acc   & Comp. \\
  \midrule
  7 Layers             & 12.4  & 37.68x   & 41.8  & 7.8x  \\
  15 Layers            & 40.9  & 18.7x    & 43.7  & 14.9x \\
  23 Layers            & 26.2  & 214.8x   & 43.5  & 7.5x  \\
  31 Layers            & 26.1  & 216.2x   & 41.8  & 5.8x  \\
  \midrule
  13 Layers*           & \textbf{44.2} & 6.2x     & \textbf{44.2} & 17.0x \\
  \bottomrule
  \end{tabular}
  \caption{Comparison of accuracy and compression rate across layers and training settings. Default settings are marked with "*".}
  \label{tab:ablation}
  \end{table}

To investigate the impact of layer selections and training strategies, we conduct comprehensive ablation studies across varying layer depths (7, 15, 23, and 31), comparing them against our primary $L=13$ layer setup. Table~\ref{tab:ablation} presents the comparative results in terms of accuracy and compression rate. Our analysis reveals two key findings: (1) Without fine-tuning, only the middle layer configuration ($L=13, 15$) achieves optimal accuracy, while others ($L=7,23,31$) perform significantly worse, supporting our hypothesis that middle layers naturally develop effective filtering mechanisms during pretraining; (2) Supervised fine-tuning substantially improves accuary and compression rate across all layer configurations, demonstrating the effectiveness of our training approach. It also indicates that the hidden states of LLMs retain rich linguistic information, which can be effectively leveraged for downstream tasks.

\subsection{Reliability of Confidence Estimates} \label{sec:confidence}% confidence 用于多轮检索
\begin{figure}[t]
  \centering
  \setlength{\abovecaptionskip}{0.cm}
  \includegraphics[width=0.98\linewidth]{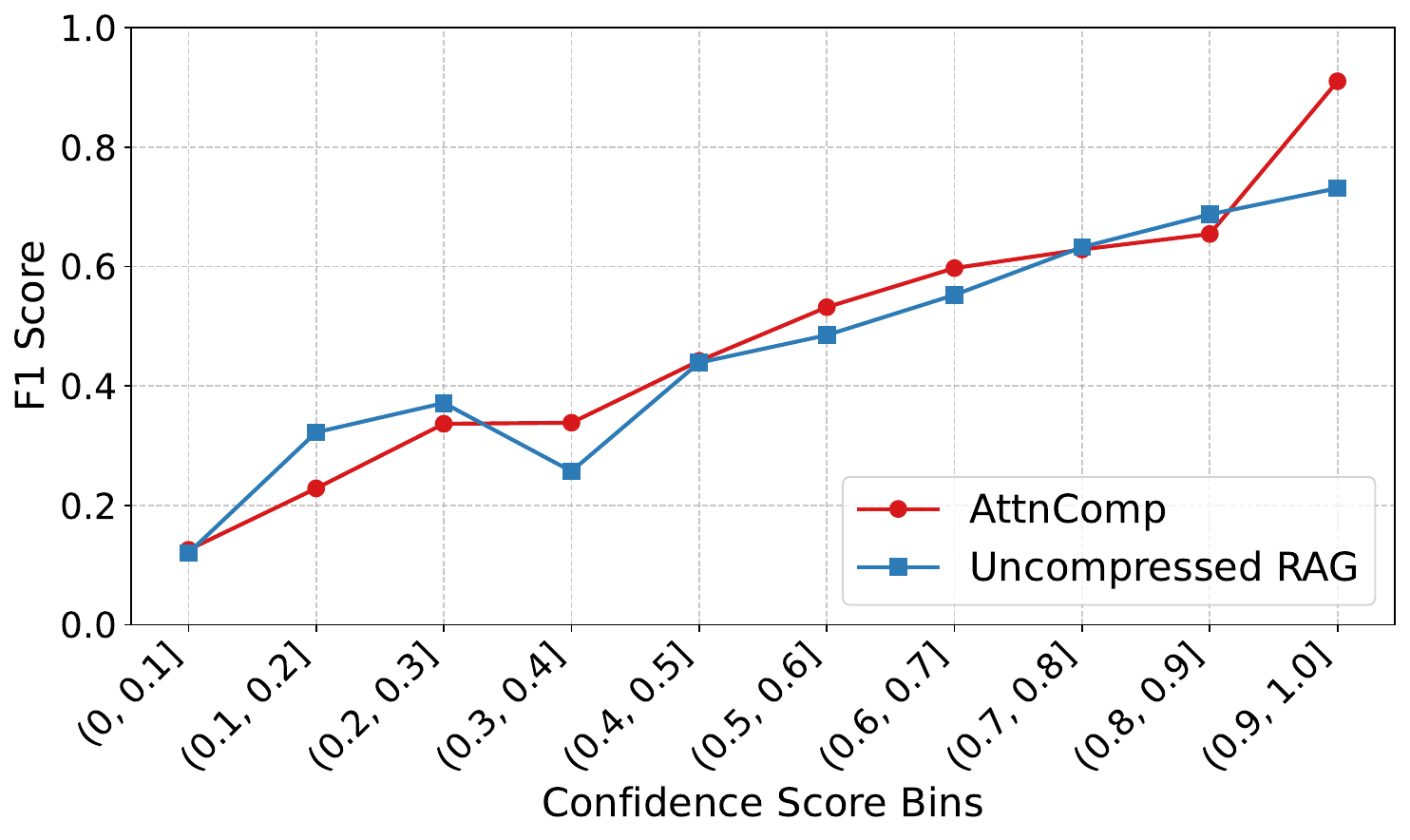}
  \caption{Average F1 score of AttnComp and Uncompressed RAG across confidence score bins on HotpotQA.}
  \label{fig:confidence}
\end{figure}

To assess the reliability of our method's confidence estimates, we compute a confidence score via Equation~\ref{eq:conf} for each test instance in the HotpotQA dataset. We then stratify these instances into ten decile groups based on their confidence scores. For each bin, we calculate the average F1 score of responses generated by both the AttnComp method and the uncompressed baseline. As shown in Figure~\ref{fig:confidence}, the results demonstrate a clear positive correlation between confidence and average F1 score for both methods. Instances with confidence scores below 0.1 yield an average F1 score of just 0.13, while those with confidence scores above 0.9 achieve a substantially higher F1 score of 0.91. Further supporting this observation, the Pearson correlation coefficient between confidence and F1 score is 0.35 for AttnComp and 0.32 for the uncompressed baseline, confirming the utility of the confidence scores as an indicator of RAG response reliability. 

We believe the confidence score can be valuable for future work on iterative RAG systems. By leveraging confidence estimates, we can assess the sufficiency of the current retrieval and set the conditions for further iterations. We leave the full exploration of such an iterative framework to future work.

%% file: appendix.tex
\section{Detailed Observations}
\label{appendix: observations}

In this section, we present detailed observations of attention behavior in LLMs, along with the corresponding experimental setup. Through three carefully designed experiments, we arrive at the following key findings:

1. Certain middle-layer attention heads effectively identify relevant information. (Section \ref{A1})

2. Attention patterns adapt to the density of relevant content. (Section \ref{A2})

3. Attention to the initial token of the context increases as the overall context becomes less relevant. (Section \ref{A3})

\subsection{Attention Heads Capture Relevance} \label{A1}
\begin{figure*}[t]
    \centering
    \setlength{\abovecaptionskip}{0.cm}
    \subfigure[Mistral-7B-v0.2]{
        \includegraphics[width=0.48\linewidth]{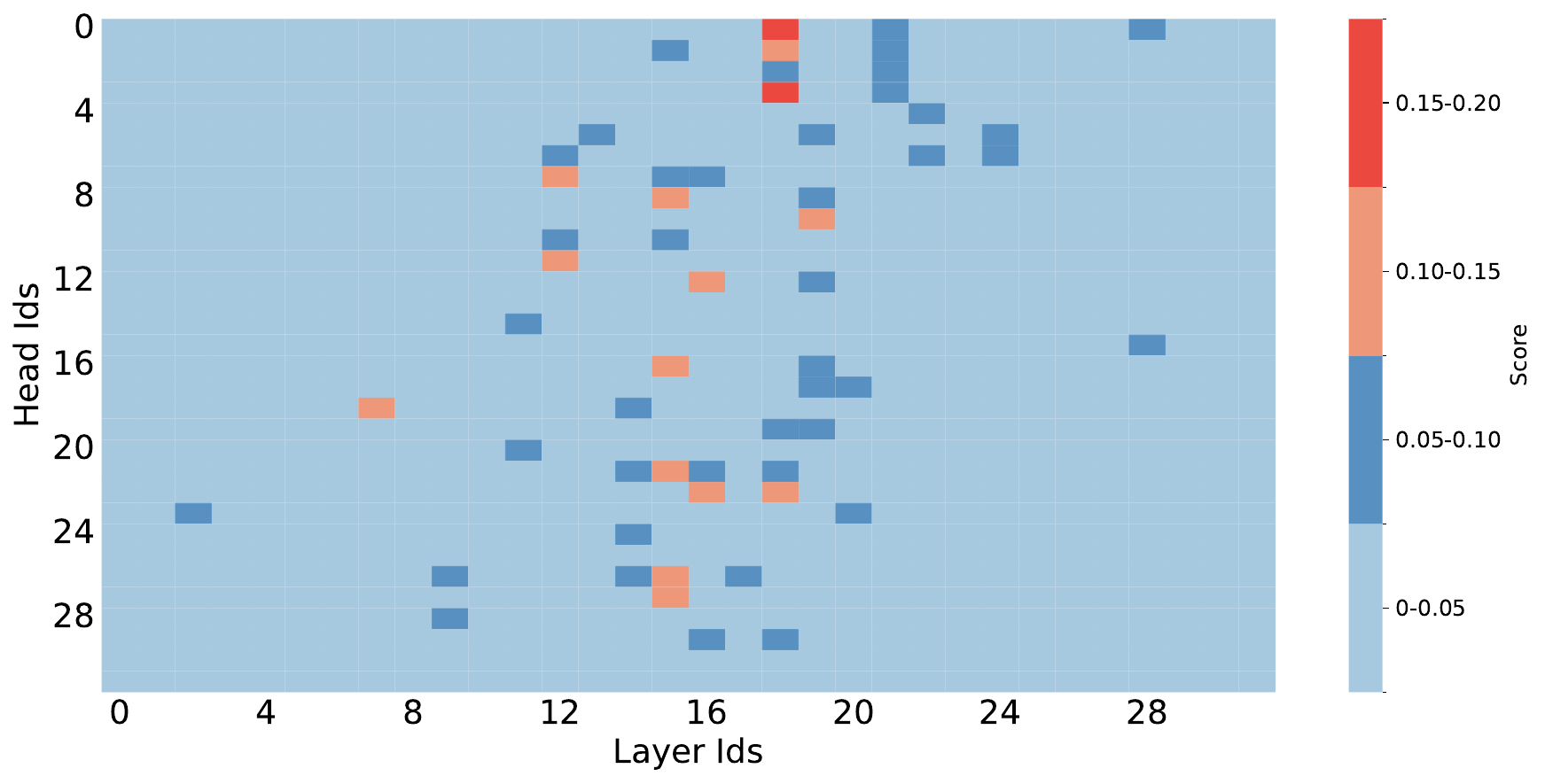}
    }
    \subfigure[Qwen2-7B-Instruct]{
        \includegraphics[width=0.48\linewidth]{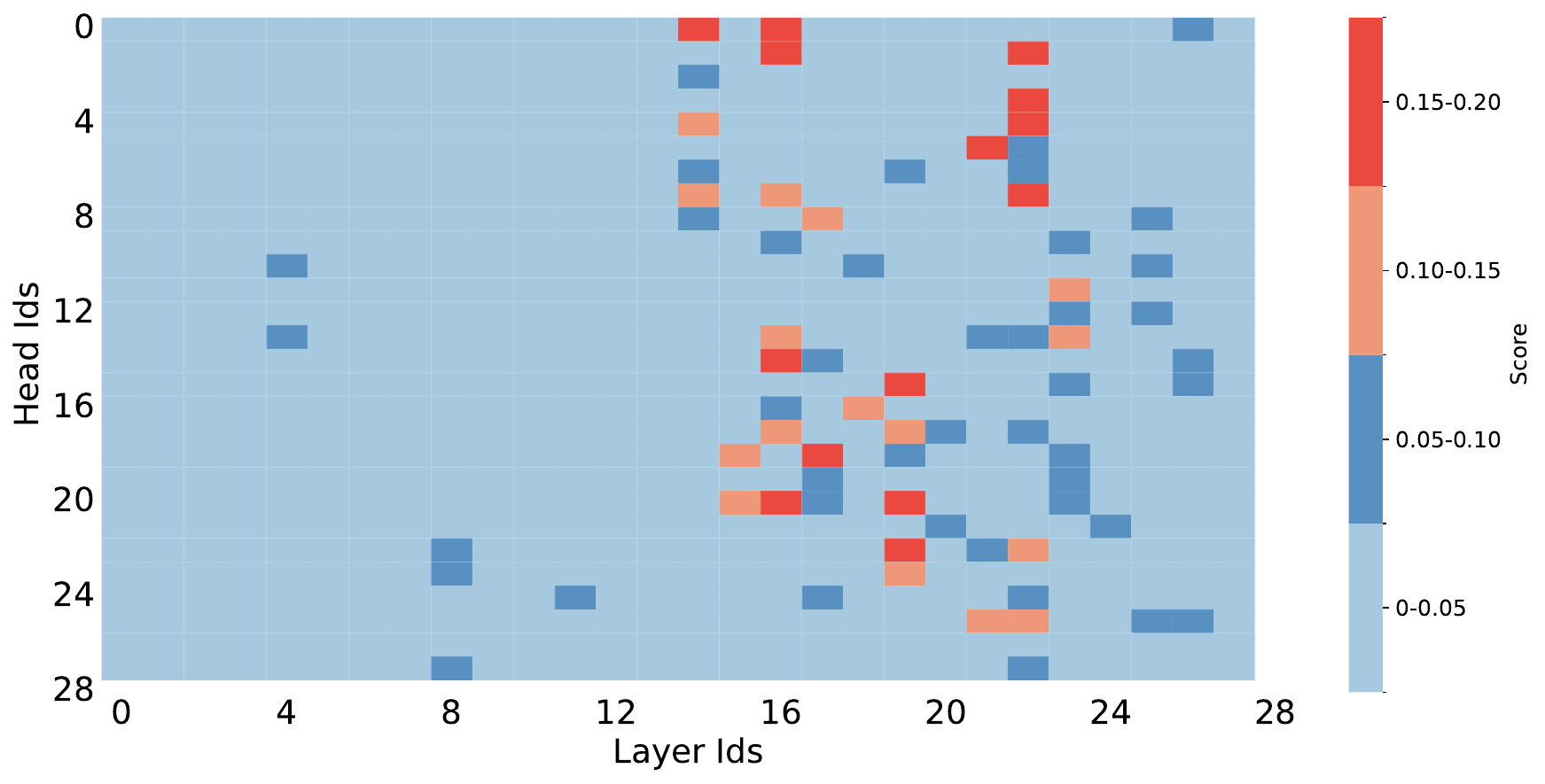}
    }
    \caption{Attention weights assigned by each attention head to supporting evidence sentences are illustrated for two LLMs: Mistral-7B-v0.2 (Figure a) and Qwen2-7B-Instruct (Figure b).}
    \label{fig:filter-heads-more}
  \end{figure*}

Prior research has revealed that LLMs exhibit retrieval heads capable of focusing on task-relevant information during text generation \citep{wu2024retrieval, fu2024not}. However, these studies primarily focus on copy-and-paste behaviors occurring during the generation phase of LLMs. Other work indicates that LLMs' attention mechanisms can identify relevant information in context before generation, yet these analyses often lack granularity—such as attention-head-level insights into how relevance is determined\citep{li2024snapkv, wu2024retrieval}. Therefore, we address the following research question: \textit{How do LLMs leverage their attention mechanisms to identify question-relevant information before text generation?}

\noindent \textbf{Experiment}
We perform our analysis on the LooGLE benchmark \citep{li2024loogle}, a long-context QA dataset where each instance consists of an article, a question, and annotated supporting evidence sentences. For each instance, we concatenate the article and question and feed them into three models—Llama-3.1-8B-Instruct \citep{grattafiori2024llama}, Mistral-7B-Instruct-v0.2 \citep{jiang2023mistral}, and Qwen2-7B-Instruct \citep{yang2024qwen2}—to compute attention weights across all attention heads. Using Equation~\ref{eq: attention-score}, we quantify the attention allocated by each head to the supporting evidence sentences, with higher scores indicating stronger focus on question-relevant information.

\noindent \textbf{Results \& Insights}
We visualize the attention scores assigned by each head to the supporting evidence sentences. As shown in Figure \ref{fig:filter-heads} for Llama-3.1-8B-Instruct, and Figure \ref{fig:filter-heads-more} for Mistral-7B-v0.2 and Qwen2-7B-Instruct, these models consistently exhibit certain middle-layer attention heads that assign noticeably higher attention to relevant evidence. In contrast, attention heads in lower and upper layers tend to show weaker focus. These findings suggest that middle-layer attention heads in LLMs are particularly effective at capturing question-relevant information within the context.

\subsection{Adaptive Attention Patterns} \label{A2}
\label{sec: Adaptive-Attention}
\begin{figure*}[t!]
    \centering
    \includegraphics[width=.98\linewidth]
    {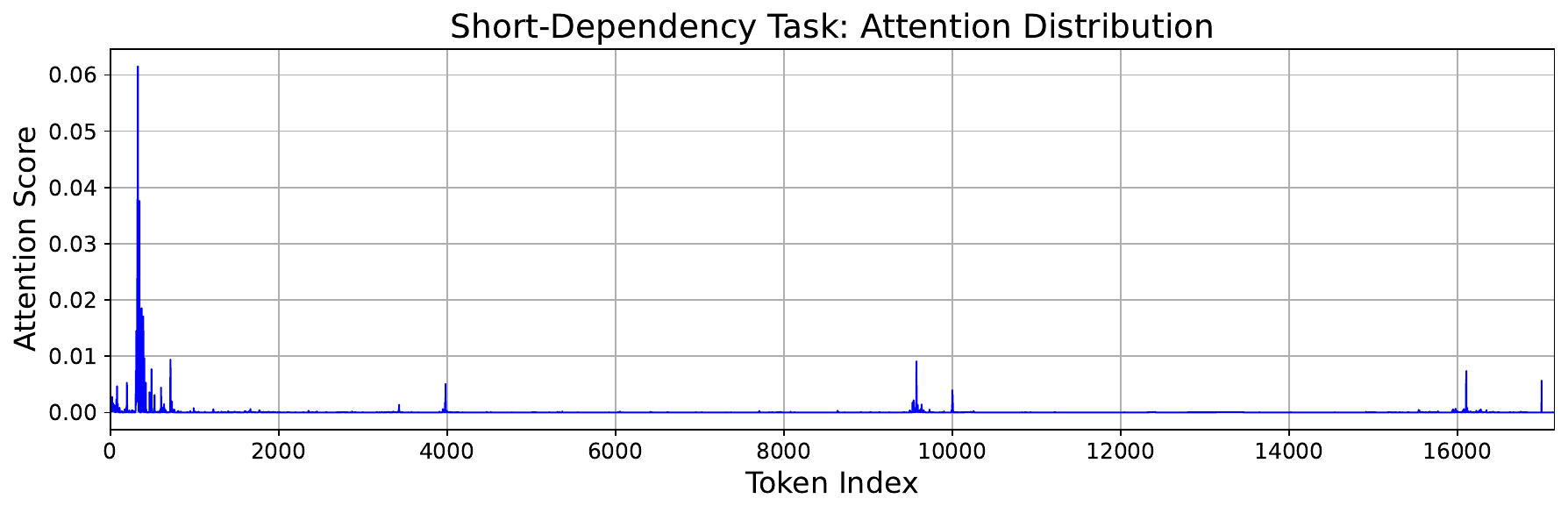}
    
    \includegraphics[width=.98\linewidth]
    {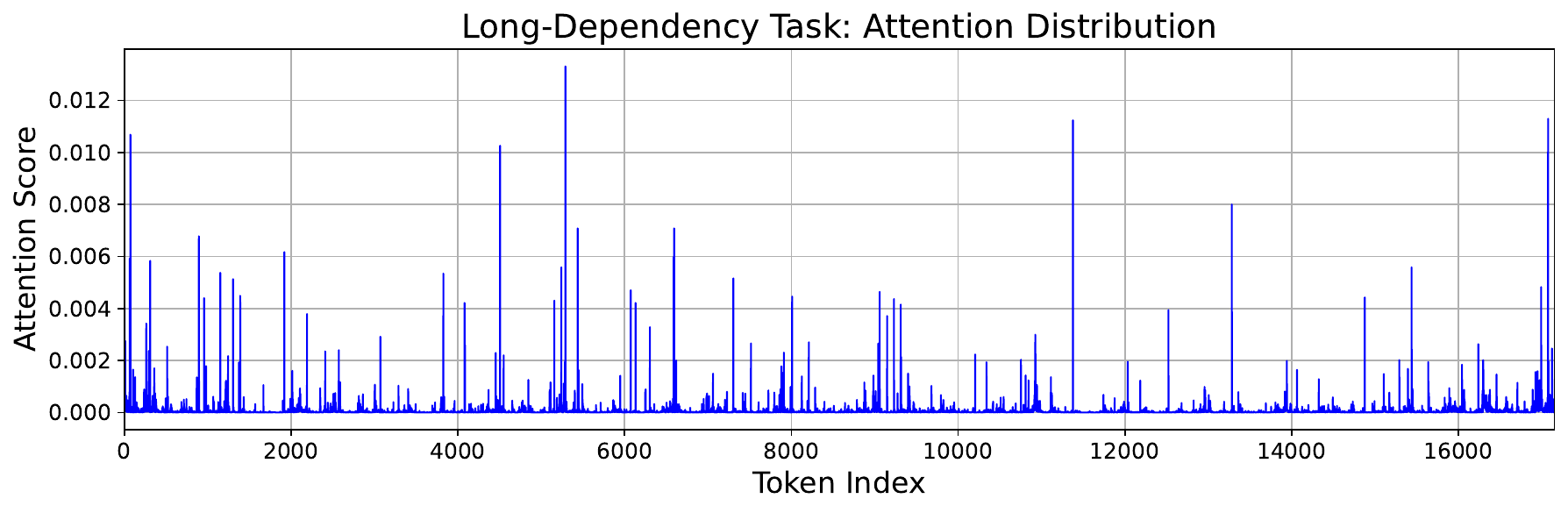}
    
    \caption{
    Examples of attention distribution for short- and long-dependency tasks. 
    Attention is more concentrated for short-dependency tasks (top), while it is more dispersed across the input for long-dependency tasks (bottom).
    }
    \label{fig: short-long attention}
\end{figure*}

The relevance of information within a context varies depending on the question and task, motivating our investigation into the model's cognitive flexibility: \textit{Does the proportion of relevant information in the context trigger distinct attention allocation strategies in LLMs?}

\noindent \textbf{Experiment} The LooGLE benchmark categorizes tasks into two types: short-dependency and long-dependency. Short-dependency tasks can be answered using a single sentence or paragraph, while long-dependency tasks require integrating information spread across multiple sentences within the article.  We sample test cases from each task category, ensuring that the input documents have a similar average number of sentences. For each sentence, we compute the attention scores and analyze the proportion of attention allocated to the top-$k$ ranked sentences. Our analysis specifically focuses on the attention heads in the 14th layer of the Llama-3.1-8B-Instruct model, as this layer has demonstrated a strong ability to capture question-relevant information.

\noindent \textbf{Results \& Insights}
Figure~\ref{fig:adaptive-attention} summarizes the experimental results. For short-dependency tasks, the attention distribution is highly concentrated, with a small number of sentences receiving a disproportionately large share of attention. In contrast, long-dependency tasks exhibit a more diffuse and uniform attention pattern across the document. Using a cumulative attention threshold of 0.8 as a reference, we observe that, on average, the top 39 sentences account for 80\% of the total attention in short-dependency cases, whereas 63 sentences are required to reach the same threshold in long-dependency tasks. This disparity suggests that the model dynamically adapts its attention mechanism based on the density and distribution of relevant information in the input context. To provide further intuition, we select representative examples from each task type and visualize their corresponding attention distributions over the context in Figure~\ref{fig: short-long attention}.

\subsection{Attention on Initial Token} \label{A3}
\label{sec: attention-sink}

In practical applications of RAG, retrieved content may sometimes be entirely irrelevant to the given question. This leads to a key research question: \textit{How do LLMs allocate attention when the retrieved context is completely irrelevant to the question?}

\noindent \textbf{Experiment} 
Inspired by prior work on attention sinks \citep{xiao2023efficient}, which reveals that initial tokens often collect significant attention scores. We hypothesize that the attention allocated to initial token increases with decreasing context relevance. To test this, we sample questions from HotpotQA \citep{yang2018hotpotqa} and PopQA \citep{mallen2023not}, and construct four context settings: (1) top 1–20 retrieved documents, (2) top 41–60 documents, (3) top 81–100 documents, and (4) 20 randomly sampled documents from the 2018 Wikipedia corpus \citep{karpukhin2020dense}. For each question, we pair it with these different context sets and measure the attention scores allocated to the initial token. The analysis is conducted using 16 attention heads from the 14th layer of the Llama-3.1-8B-Instruct model.

\noindent \textbf{Results \& Insights}
The results are shown in Figure~\ref{fig:attention-sink}. We observe a consistent increase in attention scores for the initial token as context relevance decreases. Across different retrieved document sets, the attention to the initial token remains relatively stable. However, a substantial rise is observed when completely irrelevant documents (i.e., random samples) are used as context. These findings suggest that attention on initial token may serve as a useful signal for estimating the relevance of retrieved content.

\section{Details of Data Construction} \label{appendix: dataset}
This section introduces an automated annotation pipeline based on question–answer pairs, with the detailed procedure outlined in Algorithm~\ref{alg:dataset-construction}. The pipeline consists of two stages: labeling and verification. In the labeling stage, we employ an untrained compressor to identify relevant documents. In the verification stage, we use an LLM to generate answers based on the documents selected in the previous stage and compare the generated answers with the ground truth to verify annotation correctness.

In the labeling stage, since untrained compressors are sensitive to the order of input documents, we perform multiple rounds of compression with different permutations of the document sequence. Only documents consistently retained across all rounds are labeled as relevant, while the rest are considered irrelevant. Furthermore, as untrained compressors in long-context settings may assign high attention to irrelevant content, we mitigate this issue by applying the Top-P compression algorithm with a high threshold (e.g., p=0.95) iteratively within each round. The process continues until no further reduction in document count is possible, thereby minimizing the impact of individual compression errors.

In the verification stage, the query and the selected relevant documents are provided to an LLM to generate an answer. If the generated answer matches the ground truth, the annotation is accepted. If it does not match, the LLM is prompted to generate an answer using the full set of retrieved documents. If the correct answer is obtained only with the full set, the annotation derived from the compressed set is deemed faulty and discarded. If the full set also fails to produce the correct answer, we infer that none of the retrieved documents are relevant to the query. In this case, we construct a negative example, where all documents are labeled as irrelevant. To reduce potential LLM-induced errors in negative example construction, we replace the labeled relevant documents with randomly sampled ones before labeling the entire set as irrelevant.

\begin{algorithm}[t]
\caption{Relevance Annotation}
\label{alg:dataset-construction}
\begin{algorithmic}[1]
\State \textbf{Input:} Compressor $M_c$, Generator $M_g$,  Corpus $C$, Top-P threshold $p$, number of shuffles $N$, input tuple$(q, D, a)$ where $q$ is query, $D$ is retrieved documents, $a$ is ground truth answer.
\State \textbf{Output:} Data Sample $(q, D^+, D^-)$ where $q$ is query, $D^+$ is relevant documents, $D^-$ is irrelevant documents.
% \For{$i = 1$ to $T$}
\For{$i = 1$ to $N$}
    \State Shuffle documents: $D_i \gets \text{permute}(D)$
    \While{True}
        \State Compute scores: $\{s_d\} \gets M_c(q, D_i)$
        \State $D_i' \gets \text{Top-P Compression}(\{s_d\}, p)$
        \If{$|D_i'| = |D_i|$}
        \State \textbf{break}
        \EndIf
        \State ${D}_i \gets {D}_i'$
    \EndWhile
\EndFor

\State Get relevant docs: $D^+ \gets \bigcap_{i=1}^N D_i$
\State Get irrelevant docs: $D^-\gets D \setminus D^+$
\State Generate answer: $a' \gets M_g(q, D^+)$
\If{$\text{Acc}(a', a) = 1$}
    \State \textbf{Return:} $(q, D^+, D^-)$
\Else
    \State Generate answer: $a'' \gets M_g(q, D)$
    \If{$\text{Acc}(a'', a) = 0$}
        \State $D_{\text{sample}} \gets \text{RandomSample}(C, |D^+|)$
        \State $D^- \gets D^-\cup D_{\text{sample}}$
        \State \textbf{Return:} $(q, \emptyset, D^-)$
    \Else
        \State \textbf{Return:} $\emptyset$ \Comment{Discard the sample}
    \EndIf
\EndIf
\end{algorithmic}
\end{algorithm}

\section{Implementation Details}
\label{appendix: training}
We construct query–document relevance annotations using question–answer pairs from the HotpotQA training set. For each QA pair, we initially retrieve 100 documents. We then apply Algorithm~\ref{alg:dataset-construction}, setting the number of shuffles to $N=3$ and the Top-P threshold to $p=0.95$. The untrained compressor $M_c$ is instantiated with Llama-3.1-8B-Instruct, using the attention outputs from its 14th layer. Llama-3.1-8B-Instruct also serves as the generator $M_g$ to validate the annotations.

We further adopt Llama-3.1-8B-Instruct as the backbone architecture for AttnComp. Based on the results in Figure~2(a), we retain $L=13$ transformer layers and select the top 16 attention heads from the 14th layer as the cross-attention layer. AttnComp is trained on four NVIDIA RTX 4090 GPUs (24 GB each) for 4 hours, using the Adam optimizer with a learning rate of $2 \times 10^{-4}$ and a batch size of 8. Training is conducted for 8 epochs, with the input document order shuffled in each epoch to mitigate positional bias in the attention mechanism.

\section{Baselines Details}
\label{appendix: baseline}
The details of the baseline methods are as follows:

\noindent (1) RECOMP-ext \citep{xu2023recomp} performs sentence-level semantic matching by selecting the top-$k$ sentences whose embeddings are most similar to the query. For experiments on multi-hop QA benchmarks (HotpotQA, 2WikiMultiHopQA, and MuSiQue), we use the model trained on HotpotQA, while for single-hop QA benchmarks (NQ and PopQA), we use the model trained on NQ. In all experiments, we select 50 sentences from the retrieved documents to ensure a fair comparison under a similar compression rate.

\noindent (2) LongLLMLingua \citep{jiang2024longllmlingua} removes unimportant tokens based on the perplexity scores generated by LLMs. We implement LongLLMLingua using the FlashRAG\citep{jin2024flashrag}, and set the compression ratio to 10\%.

\noindent (3) CompAct \citep{yoon2024compact} is an abstractive compression method that leverages LLMs fine-tuned on the HotpotQA dataset to generate summaries of retrieved documents. We use the publicly available implementation and model released by the authors, keeping all configurations consistent with the original setup.

\noindent (4) Provence \citep{chirkova2025provence} trains a lightweight DeBERTa model\citep{he2021debertav3} to predict sentence-level relevance scores and retains only the sentences that exceed a predefined threshold. We use the publicly available implementation and model released by the authors, keeping all configurations consistent with the original setup.

\noindent (5) AttnComp (w/o SFT) is our proposed method without supervised fine-tuning. 
We set the threshold $p$ to 0.5 to ensure a fair comparison at a similar compression rate. 

All baseline methods utilize the FlashRAG codebase \citep{jin2024flashrag} to generate answers and evaluate performance.

\section{Details of Robustness Analysis} \label{appendix: robustness}

\begin{figure*}[t]
    \centering
    \setlength{\abovecaptionskip}{0.cm}
    \subfigure[Top-k Variants Analysis]{
        \includegraphics[width=0.48\linewidth]{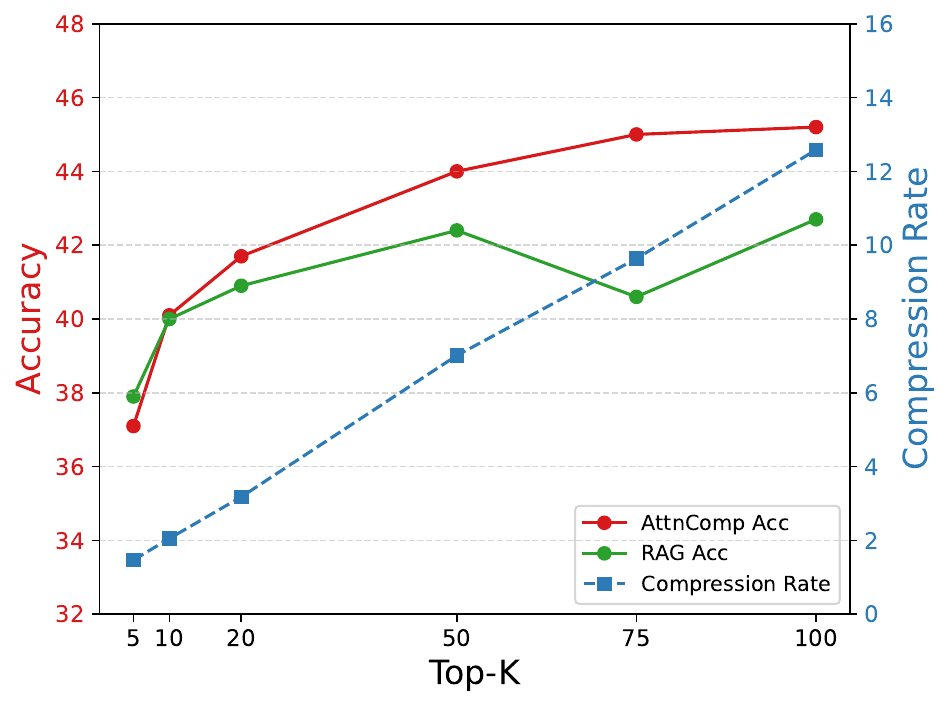}
        \label{fig:top-k}
    }
    \subfigure[Top-p Threshold Analysis]{
        \includegraphics[width=0.48\linewidth]{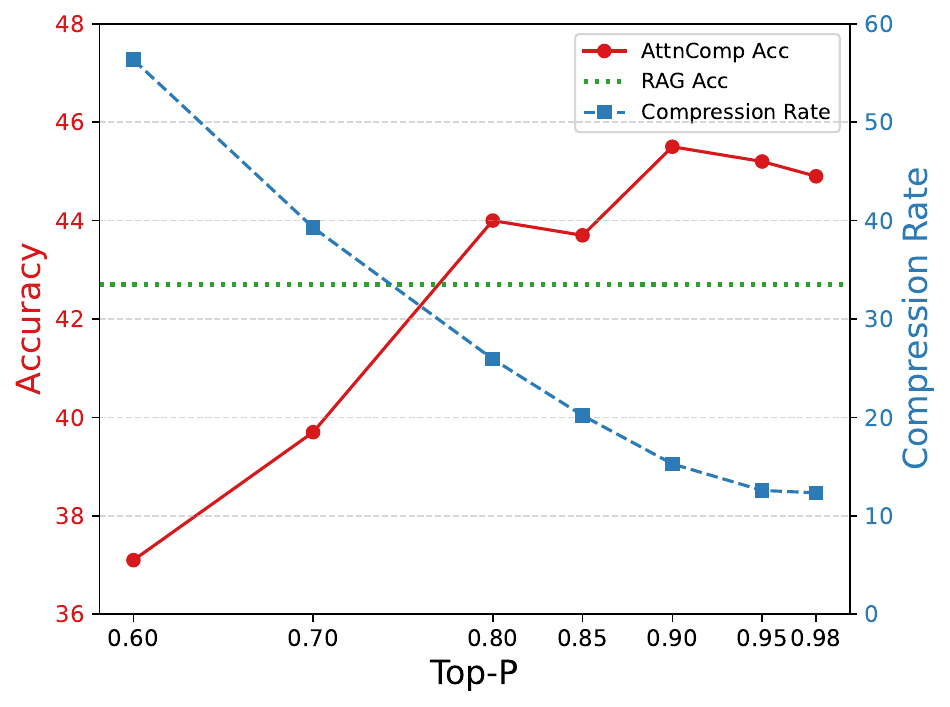}
        \label{fig:top-p}
    }
    \caption{Performance of AttnComp with varying top-k and top-p values on HotpotQA.}
    \label{fig:top-kp}
  \end{figure*}
We assess the model's robustness through experiments that vary the number of retrieved documents, top-p thresholds, and context granularities. The specific experimental settings and results are detailed as follows:

\noindent \textbf{Experiment Settings for Number of Retrieved Documents}:
We conduct experiments on HotpotQA by varying the number of retrieved documents, $k \in \{5, 10, 20, 50, 75, 100\}$, while keeping the Top-P threshold fixed at $p=0.95$. The results are presented in Figure~\ref{fig:top-k}, showing that our method consistently achieves strong performance across different values of $k$.

\noindent \textbf{Experiment Settings on Top-P Threshold}:
We conduct experiments on HotpotQA by varying the top-p threshold $p \in \{0.6, 0.7, 0.8, 0.9, 0.95, 0.98\}$, while keeping the number of retrieved documents $k$ constant at $100$.  The results in Figure~\ref{fig:top-p} demonstrate that the parameter $p$ functions as a control variable that balances accuracy against compression rate. Notably, our approach achieves consistently high performance and stability when $p$ is set to 0.9 or higher.

\noindent \textbf{Experiment Settings for Context Granularities}:
We also evaluate AttnComp under different context granularities, including document-level and sentence-level compression. For the sentence-level setting, we segment the retrieved documents into sentences following Provence \citep{chirkova2025provence}, and then apply the Top-P compression algorithm with a threshold of $p=0.95$ and a minimal score threshold of $\epsilon=10^{-3}$. As reported in Table~\ref{tab:granularity}, sentence-level compression achieves accuracy comparable to document-level compression, demonstrating that our method operates effectively across different granularities.

\begin{table}[t]
    \centering
    \small
    \begin{tabular}{c|cc|cc}
    \toprule
    \multirow{2}{*}{Dataset} & \multicolumn{2}{c|}{Document-level} & \multicolumn{2}{c}{Sentence-level} \\
    \cmidrule{2-5}
    & \textbf{Comp.} & \textbf{Acc} & \textbf{Comp.} & \textbf{Acc} \\
    \midrule
    HotpotQA   & 12.6x & 45.2 & 14.5x & 43.2 \\
    2WikiMQA   & 18.4x & 38.1 & 21.3x & 34.4 \\
    MuSiQue    & 16.3x & 19.6 & 18.5x & 20.1 \\
    NQ         & 13.5x & 53.0 & 16.8x & 51.8 \\
    PopQA      & 23.9x & 65.1 & 34.3x & 65.9 \\
    \midrule
    \textbf{Average} & 17.0x & 44.2 & 21.1x & 43.1 \\
    \bottomrule
    \end{tabular}
    \caption{Comparison of compression ratio (Comp.) and answer accuracy (Acc) between document-level and sentence-level granularity across five QA datasets using AttnComp.}
    \label{tab:granularity}
\end{table}

\section{Additional Results}
\label{appendix: additional-results}
We conduct experiments using Qwen2.5-7B-Instruct-1M \citep{yang2025qwen2} as the reader model, which exhibits stronger long-context capabilities. All other experimental settings remain consistent with the main experiments. The results, summarized in Table~\ref{table:results-qwen}, show that AttnComp consistently outperforms other compression approaches across all datasets while achieving a high compression rate. Notably, our approach even surpasses the uncompressed baseline in terms of F1. These results further validate the effectiveness of AttnComp.
\begin{table*}[]
  \centering
  \resizebox{1.0\textwidth}{!}{
  \begin{tabular}{l|ccc|ccc|ccc|ccc|ccc|ccc}
  \toprule
    {\multirow{2}{*}{\textbf{Methods}}} &
     \multicolumn{3}{c|}{\textbf{HotpotQA}} &
     \multicolumn{3}{c|}{\textbf{2WikiMQA}} &
     \multicolumn{3}{c|}{\textbf{MuSiQue}} &
     \multicolumn{3}{c|}{\textbf{NQ}} &
     \multicolumn{3}{c|}{\textbf{PopQA}} &
     \multicolumn{3}{c}{\textbf{AVG}} \\ \cmidrule{2-19} 
    {}     &     {\textbf{Comp.}} &   {\textbf{F1}}   &   {\textbf{Acc}}   &   {\textbf{Comp.}} &   {\textbf{F1}}   &   {\textbf{Acc}}  &    {\textbf{Comp.}} &   {\textbf{F1}}   &   {\textbf{Acc}} &    {\textbf{Comp.}} &   {\textbf{F1}}   &   {\textbf{Acc}} &    {\textbf{Comp.}} &   {\textbf{F1}}   &   {\textbf{Acc}} & {\textbf{Comp.}} &   {\textbf{F1}}   &   {\textbf{Acc}} \\
    
  \midrule
  \multicolumn{1}{l}{ \textbf{\textit{No Retrieval}}} \\
  {Direct}     
    & - &  26.5 & 23.6
    & - & 26.2 &  34.1
    & - & 11.4 & 8.8
    & - & 27.1 & 23.6
    & - & 24.1 & 31.3
    & - & 23.1 & 24.3 \\
  \midrule
  \multicolumn{1}{l}{\textbf{\textit{Retrieval without Compression}}} \\
  {All Documents}    
  & {1x}  & {46.4} & {42.3}
  & {1x} & {38.5} & \textbf{38.8}
  & {1x} & {22.4} & \textbf{20.0}
  & {1x} & {42.6} & {51.1}
  & {1x} & {27.5} & \textbf{65.0}
  & 1x & 35.5 & 43.4 \\

  {Top 5 Documents}
  & {18.2x}  & {42.3} & {36.5} 
  & {18.3x}  & {35.4} & {33.1} 
  & {18.4x}  & {17.9} & {15.0}
  & {18.3x} & {48.5} & {51.3} 
  & {18.4x} & {37.4} & {60.2}
  & 18.3x & 36.3 & 39.2 \\
  
  {Top 10 Documents}
  & {9.6x}  & {44.3} & {38.5}
  & {9.6x}  & {38.4} & {36.8}
  & {9.6x}  & {19.8} & {16.9}
  & {9.6x} & \textbf{49.0} & \textbf{52.7}
  & {9.6x} & {37.5} & {62.2}
  & {9.6x} & 37.8 & 41.4 \\
  \midrule
  \multicolumn{1}{l}{\textbf{\textit{Retrieval with Compression}}} \\
  {RECOMP-ext}
  & {8.0x}  & {40.6} & {35.5}
  & {8.0x}  & {37.5} & {35.1}
  & {8.1x}  & {19.9} & {16.3}
  & {8.4x}  & {41.8} & {44.8}
  & {9.0x} & {29.9} & {51.0}
  & 8.3x & 33.9 & 36.5  \\ 
  
  {LongLLMLingua}
  & {9.7x} & {45.1} & {39.5}
  & {9.7x} & {35.7} & {33.2}
  & {9.7x} & {18.2} & {14.1}
  & {9.7x} & {39.0} & {41.5}
  & {9.7x} & {33.4} & {58.6}
  & 9.7x & 34.3 & 37.4 \\ 
  
  {CompAct} 
  & {80.0x} & {45.8} & {40.4}
  & {82.4x} & {34.9} & {35.7}
  & {71.8x} & {18.0} & {16.6}
  & {84.2x} & {44.7} & {47.7}
  & {98.0x} & {39.7} & {57.4}
  & {83.3x} & 36.6 & 39.6 \\
  
  {Provence} 
  & {10.2x} & {44.3} & {39.4}
  & {10.7x} & {33.7} & {31.4}
  & {8.7x}  & {19.5} & {16.6}
  & {6.8x}  & {43.2} & {46.8}
  & {6.9x} & {30.2} & {55.7}
  & 8.7x & 34.2 & 38.0 \\ 
  \midrule
  \rowcolor{blue!10} {AttnComp (Ours)} 
  & {12.6x} & \textbf{50.8} & \textbf{45.4}
  & {18.4x} & \textbf{40.5} & {38.1}
  & {16.3x} & \textbf{23.4} & {19.6}
  & {13.5x} & {48.4} & {50.1}
  & {23.9x} & {39.8} & {63.9}
  & 17.0x & \textbf{40.6} & \textbf{43.4} \\ 

  \bottomrule
  \end{tabular}}
  \caption{Results with Qwen2.5-7B-Instruct-1M~\citep{yang2025qwen2} as the reader model; all other experimental settings are kept the same as in the main results.}
  \label{table:results-qwen}
%   \vspace{-0.3cm}
\end{table*}

\section{Case Study} \label{appendix: case-study}
In Table \ref{table:case}, we present a representative example from the HotpotQA dataset. The query is: "\textit{Who was the eldest brother of the Mexican drug trafficker born 12 March 1952?}" Two of retrieved documents provide the necessary evidence. Document A states, "\textit{Benjamín Arellano Félix (born 12 March 1952) is a Mexican drug trafficker}" (see the first document in the AttnComp compressed context), while Document B indicate that, "\textit{Francisco Rafael Arellano Félix is the eldest brother of Benjamín Arellano Félix}" (see the third document in the AttnComp compressed context). Importantly, the relevance of Document B is not evident in isolation, as it requires the contextual link provided by Document A. Without this cross-document connection, Document B is prone to being mistakenly filtered out as irrelevant.

AttnComp addresses this issue by jointly processing all retrieved documents, allowing it to capture semantic dependencies across documents and retain both supporting facts. In contrast, methods such as RECOMP\citep{xu2023recomp}, LongLLMLingua\citep{jiang2024longllmlingua} and Provence\citep{chirkova2025provence} process each document independently, preventing them from integrating cross-document information and often leading to the erroneous exclusion of relevant content. Although CompAct\citep{yoon2024compact} adopts an iterative integration mechanism, it often halts the iteration prematurely before gathering sufficient evidence, ultimately missing the key facts needed to answer the query.

% Define custom highlight colors
\colorlet{hlgreen}{green!30}
\colorlet{hlred}{red!30}

\newcommand{\hlgreen}[1]{{\sethlcolor{hlgreen}\hl{#1}}}
\newcommand{\hlred}[1]{{\sethlcolor{hlred}\hl{#1}}}
\begin{table*}[t]
\footnotesize
\centering{
\begin{tabular}{p{0.95\textwidth}}
\toprule
\textbf{Question}: Who was the eldest brother of the Mexican drug trafficker born 12 March 1952? \\
\textbf{Answer:} Francisco Rafael Arellano Félix \\
\midrule
\textbf{Method}: \textsc{AttnComp} (Ours) \\
\textbf{Compressed Context}: \\
Doc 1(Title: "Benjamín Arellano Félix") \hlgreen{Benjamín Arellano Félix (born 12 March 1952) is a Mexican drug trafficker} and former leader of the Mexican criminal organization known as the Tijuana Cartel or ""Arellano-Félix Organization"". Benjamín Arellano Félix, who worked closely with his brothers, was one of Mexico's most powerful drug lords and the supplier of one-third of the U.S.'s cocaine. Benjamín had six brothers: He also has four sisters. Two of them, Alicia

\dots

Doc 3(Title: "Francisco Rafael Arellano Félix") \hlgreen{Francisco Rafael Arellano Félix Francisco Rafael Arellano Félix (24 October 1949 – 18 October 2013) was a Mexican drug lord and former leader of the Tijuana Cartel, a drug trafficking organization. He was the oldest of seven brothers and headed the criminal organization early in the 1990s alongside them. Through his brother Benjamín}, Francisco Rafael joined the Tijuana Cartel in 1989 following the arrest of Miguel Ángel Félix Gallardo
\\
\textbf{Predict}: 
\hlgreen{Francisco Rafael Arellano Félix} (\textbf{Correct}) \\ 

\midrule
\textbf{Method}: \textsc{Recomp} \\
\textbf{Compressed Context}: \\
(Title: "Eduardo Arellano Félix") Eduardo Arellano Félix Eduardo Arellano Félix (born October 11, 1956) is a Mexican drug trafficker, brother of Benjamín, Ramón, Javier and sister Enedina, all drug traffickers.

\dots

(Title: "Jorge Luis Ochoa Vásquez") \hlred{Jorge Luis Ochoa Vásquez Jorge Luis Ochoa Vásquez (September 30, 1950) is a Colombian drug trafficker who was one of the key founding members of the notorious Medellín Cartel in the late 1970s.}

\dots

(Title: "Ramón Arellano Félix") Ramón Arellano Félix Ramon Arellano Félix (August 31, 1964 – February 10, 2002) was a Mexican drug trafficker whom authorities linked to the Tijuana drug cartel

\textbf{Predict}: 
\hlred{Jorge Luis Ochoa Vásquez} (\textbf{Wrong}) \\ 

\midrule
\textbf{Method}: \textsc{LongLLMLingua} \\
\textbf{Compressed Context}: \\
Doc(Title: "amín Arellano Félix display at Museo del Enervante Mexico City. currently incarcerated at United States Penitentiary Canaan In the217 Netflix andivision series,El Chapo"", Hern Rom Benjam Avendañoa fictionalized portrayal of Benjamín Arellano Félix)

\dots

\hlgreen{Doc 2(Title: "Benjamín Arellano Féli Benjamín Arellano Félix Benjamín Arellano Féli (born 12 March 1952) is a Mexican drug traff} and former leader of Mexican criminal organization known as the Tijuana Cartel or ""Arellano-Félix Organization" Benjamín Arellano Féli, who worked closely with his brothers, was one of Mexico's most powerful lords andlier of one-third of the U..'s cocaine. Benjamín had brothers: also has four sisters....

\dots

He has several aliases, including El 85, Saúl Ulloa Cuevas, Gerardo Sánchez Espinosa, Érick Valencia Cornelio, Ochenta y Cinco, and Mono. His criminal profile says he is tall and weighs

\textbf{Predict}: 
\hlred{This information is not available in the given documents} (\textbf{Wrong}) \\ 

\midrule
\textbf{Method}: \textsc{CompAct} \\
\textbf{Compressed Context}: \\
\hlgreen{Benjamín Arellano Félix, born on 12 March 1952, is a Mexican drug trafficker} and former leader of the Mexican criminal organization known as the Tijuana Cartel or the Arellano-Félix Organization. Benjamín had six brothers, including He also has four sisters.\\
\textbf{Predict}: 
\hlred{Benjamín Arellano Félix} (\textbf{Wrong}) \\ 

\midrule
\textbf{Method}: \textsc{Provence} \\
\textbf{Compressed Context}: \\
(Title: "Eduardo Arellano Félix") Eduardo Arellano Félix Eduardo Arellano Félix (born October 11, 1956) is a Mexican drug trafficker, brother of Benjamín, Ramón, Javier and sister Enedina, all drug traffickers.

\dots

\hlred{Juan David was the elder brother of Jorge Luis and Fabio Ochoa Vásquez, powerful figures inside
Born in a small town in the state of Sinaloa, Torres Félix began working for the Sinaloa Cartel in the 1990s and later ascended to the apex of the cartel after his brother Javier Torres Félix was arrested in 2004.}
He reportedly has five brothers: Nemesio, Juan, Miguel, Marín, and Abraham.

\dots

(Title: "Enedina Arellano Félix") brother Eduardo Arellano Félix in 2008.
Benjamín Arellano Félix, who worked closely with his brothers, was one of Mexico's most powerful drug
He formed the Beltrán Leyva Cartel along with his brothers Héctor, Carlos and Arturo.

\textbf{Predict}: 
\hlred{Juan David Ochoa Vásquez} (\textbf{Wrong}) \\

\bottomrule
\end{tabular}
}

\caption{Case study comparing compressed contexts and answers generated by baseline methods and AttnComp. Relevant content and correct answers are highlighted in green, while misleading content and incorrect answers are highlighted in red.}
\label{table:case}
\end{table*}